%% file: main.tex
\documentclass[10pt,letterpaper]{article}
\usepackage[pagenumbers]{cvpr} 


\usepackage[dvipsnames,table]{xcolor} 
\definecolor{cvprblue}{rgb}{0.21,0.49,0.74}
\definecolor{lightpurple}{rgb}{0.902, 0.863, 0.922}
\definecolor{lightgreen}{rgb}{0.863, 0.902, 0.863}
\definecolor{lightpink}{rgb}{0.980, 0.941, 0.941}

\usepackage[pagebackref,breaklinks,colorlinks,allcolors=cvprblue]{hyperref}
\usepackage{amssymb}
\usepackage{multirow}
\usepackage{makecell}
\usepackage{float}

\graphicspath{{./fig/}}

\newcommand{\customtitle}[2]{%
  \begin{center}
    {\large \textbf{#1}} \\[0.5em] 
    \vspace{1em} 
  \end{center}
}

\makeatletter
\title{Pixel-level and Semantic-level Adjustable Super-resolution: \\ A Dual-LoRA Approach}
\author{Lingchen Sun$^{1,2}$,  Rongyuan Wu$^{1,2}$,  Zhiyuan Ma$^{1}$, Shuaizheng Liu$^{1,2}$, Qiaosi Yi$^{1,2}$, Lei Zhang$^{1,2}$\thanks{Corresponding author. This research is supported by the PolyU-OPPO Joint Innovative Research Center.} \\
{$^{1}$The Hong Kong Polytechnic University \qquad $^{2}$OPPO Research Institute}
}

\begin{document}
\twocolumn
\maketitle

\input{sec/0_abstract}    
\input{sec/1_intro}

\input{sec/2_Related_Work}
\input{sec/3_Method}
\input{sec/4_Experiement}
\input{sec/5_conclusion}

{
    \small
    \bibliographystyle{ieeenat_fullname}
    \bibliography{main}
}

\newpage
\onecolumn

\customtitle{Supplementary Material to ``Pixel-level and Semantic-level Adjustable Super-resolution: A Dual-LoRA Approach''}

\input{sec/supplementary}

\end{document}

%% file: sec/0_abstract.tex
\begin{abstract}
Diffusion prior-based methods have shown impressive results in real-world image super-resolution (SR). However, most existing methods entangle pixel-level and semantic-level SR objectives in the training process, struggling to balance pixel-wise fidelity and perceptual quality. Meanwhile, users have varying preferences on SR results, thus it is demanded to develop an adjustable SR model that can be tailored to different fidelity-perception preferences during inference without re-training. 
We present \textbf{Pi}xel-level and \textbf{S}emantic-level \textbf{A}djustable \textbf{SR} (\textbf{PiSA-SR}), which learns two LoRA modules upon the pre-trained stable-diffusion (SD) model to achieve improved and adjustable SR results.
We first formulate the SD-based SR problem as learning the residual between the low-quality input and the high-quality output, then show that the learning objective can be decoupled into two distinct LoRA weight spaces: one is characterized by the $\ell_2$-loss for pixel-level regression, and another is characterized by the LPIPS and classifier score distillation losses to extract semantic information from pre-trained classification and SD models. 
In its default setting, PiSA-SR can be performed in a single diffusion step, achieving leading real-world SR results in both quality and efficiency. 
By introducing two adjustable guidance scales on the two LoRA modules to control the strengths of pixel-wise fidelity and semantic-level details during inference, PiSA-SR can offer flexible SR results according to user preference without re-training.
The source code of our method can be found at \href{https://github.com/csslc/PiSA-SR}{https://github.com/csslc/PiSA-SR}. 

\end{abstract}

%% file: sec/1_intro.tex
\section{Introduction}
\vspace{-0.3em}
\label{sec:Intro}
\begin{figure}
	\centering 
	\includegraphics[scale=0.42]{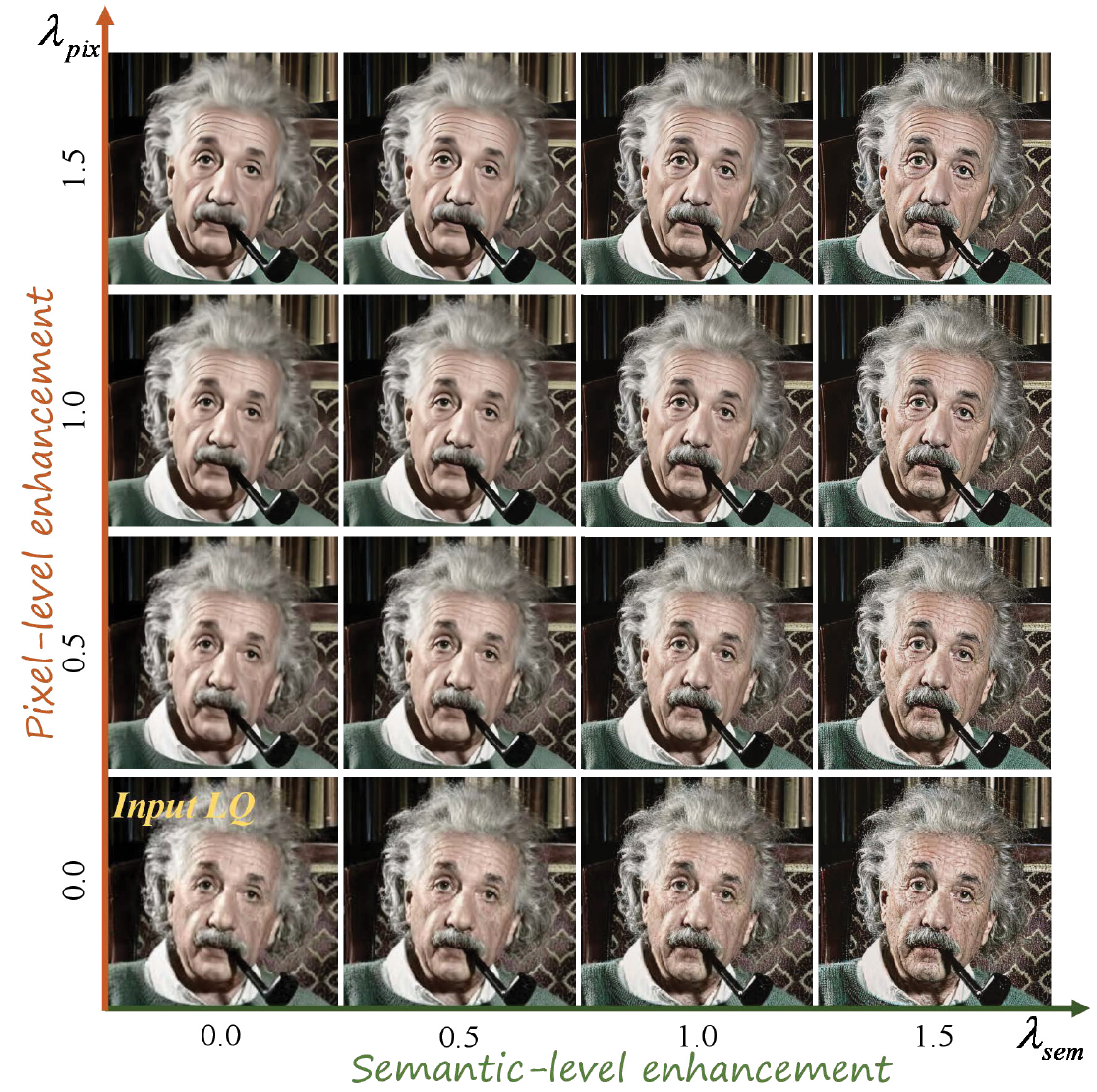}
	\vspace{-1.em}
	\caption{Visual illustration of our pixel- and semantic-level adjustable method for real-world SR. By increasing the pixel-level guidance scale $\lambda_{pix}$, the image degradations such as noise and compression artifacts can be gradually removed; however, a too-strong $\lambda_{pix}$  will make the SR image over-smoothed. By increasing the semantic-level guidance scale $\lambda_{sem}$, the SR images will have more semantic details; nonetheless, a too-high $\lambda_{sem}$ will generate visual artifacts. Please zoom in for a better view.}
	\label{fig1}
 \vspace{-1.em}
\end{figure}

Single image super-resolution (SR) \cite{9044873} aims to reconstruct a high-quality (HQ) image from its low-quality (LQ) counterpart suffering from various degradations, \eg, noise, blur, downsampling, \etc. SR is a challenging ill-posed problem due to the inherent ambiguity in reconstructing details \cite{pdtradeoff, LDL}, and various deep neural networks (DNNs) \cite{srcnn, haris2018deep, kim2016deeply, swinir, elan, hat2023} have been developed to address it. The pixel-level regression losses, such as $\ell_1$ and $\ell_2$ losses, are essential to keep the pixel-level fidelity of SR outputs, yet they tend to produce over-smoothed details \cite{pdtradeoff, LDL}.
The SSIM \cite{ssim} and perceptual losses can alleviate this problem. In specific, SSIM measures the local structural similarity between the SR and ground-truth (GT) images, and the perceptual loss \cite{perceptualloss} extracts semantic features with a pre-trained classification model \cite{vgg} to improve the perceptual quality of SR results. 
The adversarial loss associated with the generative adversarial network (GAN) \cite{srgan,esrgan, realesrgan, LDL, DASR} provides a more effective solution than SSIM/perceptual losses to align SR images with natural image distribution, thereby leading to perceptually more realistic SR results. However, it simultaneously introduces many undesirable visual artifacts \cite{LDL} because the adversarial training is unstable and the limited capacity of GAN models in characterizing the space of natural images.

Recently developed large-scale stable diffusion (SD) models \cite{stablediffusion}, which are pre-trained on text-to-image (T2I) tasks, have demonstrated impressive semantic understanding and have been successfully applied to many downstream tasks \cite{controlnet,t2i}, including SR \cite{stablesr,diffbir,pasd, seesr,osediff, ccsr}. While SD-based SR approaches have exhibited perceptually more realistic SR results than GAN-based methods \cite{chen2025generalizedimagequalityassessment}, they commonly entangle the pixel-level fidelity and semantic-level enhancement objectives in the diffusion process \cite{stablesr,seesr,osediff}, which can be contradictory in optimization \cite{pdtradeoff, eaadam}. As a result, they struggle to balance the pixel-wise fidelity and semantic-level perception in the final SR output. While some methods \cite{diffbir, pasd, ccsr} perform pixel-level restoration before semantic-level enhancement, the performance of the latter stage depends heavily on the accuracy of the earlier stage. 

Besides the trade-off between pixel-level fidelity and semantic -level details, users in practice often have different preferences on the SR results: some prioritize content fidelity over detail generation, while others favor semantically rich details over pixel-wise fidelity.
This diversity in user preferences underscores the importance and demands for a more flexible SR approach that can accommodate individual tastes during inference.
Although some previous methods \cite{he2021interactive,mou2022metric} have explored interactive SR methods to control the restoration strength, they are primarily limited to degradation-level adjustments (such as noise and blur), resulting in either smoother or sharper outputs.
Several multi-step SD-based SR methods \cite{seesr, pasd, diffbir,supir} employ guidance-based strategies \cite{cfg} at each sampling step to achieve varying levels of semantic richness, but these methods often struggle with precise control and efficiency.

In this paper, we present \textbf{Pi}xel-level and \textbf{S}emantic-level \textbf{A}djustable \textbf{S}uper-\textbf{R}esolution, namely \textbf{PiSA-SR}, which separates pixel-level and semantic-level enhancement into two distinct Low-Rank Adapter (LoRA) \cite{hu2021lora} weight spaces with the pre-trained SD model, providing an effective SR model for various user preferences. 
 We begin by formulating the SD-based SR problem to learn the residual between LQ and HQ latent features \cite{resnet,dncnn,zhang2018residual}. With this formulation, we can not only accelerate the convergence of model training but also introduce scale factors on the model outputs, enabling flexible adjustment on the SR results during the inference stage without re-training.
Then, we fine-tune the pre-trained SD model by introducing two separate LoRA modules dedicated to pixel-level regression and semantic-level enhancement. In specific, the $\ell_2$ loss is employed for the LoRA module of pixel-level regression, while the LPIPS \cite{lpips} and classifier score distillation (CSD) \cite{csd} losses are employed for another LoRA module, leveraging the semantic priors encoded in the pre-trained VGG classification model \cite{vgg} and SD image generation model \cite{stablediffusion}.
A decoupled training approach is proposed to train pixel- and semantic-level LoRAs, effectively improving semantic information while maintaining pixel-level fidelity.

Our experiments demonstrate that PiSA-SR not only achieves superior SR performance to existing SD-based models but also offers an effective way for users to adjust the SR styles based on their tastes. An example is shown in Fig. \ref{fig1}. 
The horizontal and vertical axes indicate the semantic-level and pixel-level guidance scales, respectively. Increasing the pixel-level scale removes noise and compression artifacts, but a too-strong scale over-smoothes image details. Conversely, increasing the semantic-level scale enriches the image details. Nonetheless, a too-high semantic scale over-enhances visual artifacts. PiSA-SR provides the flexibility to adjust both pixel-level and semantic-level guidance scales based on user preferences.

%% file: sec/2_Related_Work.tex
\section{Related Work}
\label{sec:Relatedwork}
\vspace{-0.3em}
\textbf{Image super-resolution}.
Early deep learning-based SR methods \cite{srcnn, haris2018deep, kim2016deeply, swinir} aim to improve image fidelity metrics such as PSNR and SSIM \cite{ssim}. By using bicubic downsampling to generate LQ-HQ image training pairs, various approaches have been developed to enhance the SR performance, such as dense \cite{denseSR}, residual \cite{zhang2018residual}, recursive connections \cite{kim2016deeply}, non-local strategies \cite{wang2021lightweight}, and attention mechanisms \cite {swinir, elan, hat2023}.
Though these advancements have significantly improved the SR performance, the real-world LQ images with complex degradations are much more challenging to enhance. While researchers have proposed to collect real-world LQ-HQ paired datasets \cite{realsr, drealsr} to train SR models, simulating realistic LQ-HQ pairs is a more economic way \cite{bsrgan,realesrgan}. BSRGAN \cite{bsrgan} randomly shuffles some basic degradation operators to synthesize LQ-HQ pairs, while RealESRGAN \cite{realesrgan} implements high-order degradation modeling.
Compared with pixel-level losses, the perceptual loss \cite{perceptualloss} and GAN loss \cite{srgan,esrgan} are more effective in improving image perceptual quality. However, the adversarial training also leads to unnatural visual artifacts in the SR results.
Many of the later works aim to reduce GAN-generated artifacts from the perspectives of frequency division \cite{bestbeddy}, network optimization \cite{zhang2022perception}, the weighting of different losses \cite{wang2019deep, eaadam}, and image local statistics \cite{LDL}. 

\textbf{Diffusion model based super-resolution}.
Recent studies have explored the use of diffusion models (DMs) \cite{ddpm, stablediffusion} for SR. Early approaches adjust pre-trained DMs' reverse transitions \cite{ddrm} using gradient descent, offering training-free solutions by assuming predefined image degradations. ResShift \cite{resshift} trains a DM from scratch on paired LQ-HQ data. 
Recently, SD models pre-trained on T2I tasks have been widely used for SR tasks due to their strong image priors \cite{stablesr,diffbir,pasd,seesr,osediff,sinsr,mei2023conditional, dreamclear,xpsr,chen2024adversarial}.
StableSR \cite{stablesr} introduces a trainable encoder and takes LQ image as a condition in SD. DiffBIR \cite{diffbir} first applies a restoration module to mitigate degradations, then utilizes SD to enhance details. 
PASD \cite{pasd} employs an encoder for degradation removal and introduces a pixel-aware cross-attention module to incorporate low-level and high-level image features into the SD process. SeeSR \cite{seesr} enhances semantic robustness through degradation-aware tag-style prompts to guide diffusion. 
Considering that the multi-step diffusion process will increase the computational cost and the risk of synthesizing unfaithful contents, researchers started to develop fewer-step DM-based SR models. 
SinSR \cite{sinsr} applies consistency-preserving distillation to shorten the diffusion process of ResShift. 
OSEDiff \cite{osediff} employs the LQ image as direct input, eliminating the random noise sampling. It adopts the VSD loss \cite{vsd, dmd} to distill the generative capacity from the multi-step diffusion process, offering a one-step DM-based SR solution. 
However, all the above-mentioned methods face a contradiction between pixel-level and semantic-level enhancement and most of them lack the flexibility to adjust the SR style for diverse user preferences. 

%% file: sec/3_Method.tex
\section{Methodology}
\vspace{-0.3em}
\label{sec:Method}
This section first formulates the SD-based SR as a residual learning model, then introduces the dual-LoRA approach to disentangle the learning objectives of pixel-level regression and semantic-level enhancement. Finally, pixel and semantic guidances are proposed for flexible SR results generation. 
In our following development, we denote by $x_L$ and $x_H$ the LQ and HQ images, respectively, and by $z_L$ and $z_H$ their corresponding latent codes. Let $E$ and $D$ be the encoder and decoder of a well-trained variational auto-encoder (VAE), we can nearly have $z_L=E(x_L)$, $z_H = E(x_H)$, $x_L=D(z_L)$ and $x_H = D(z_H)$.

\subsection{Model Formulation}
\begin{figure}
	\centering 
    	\includegraphics[scale=0.43]{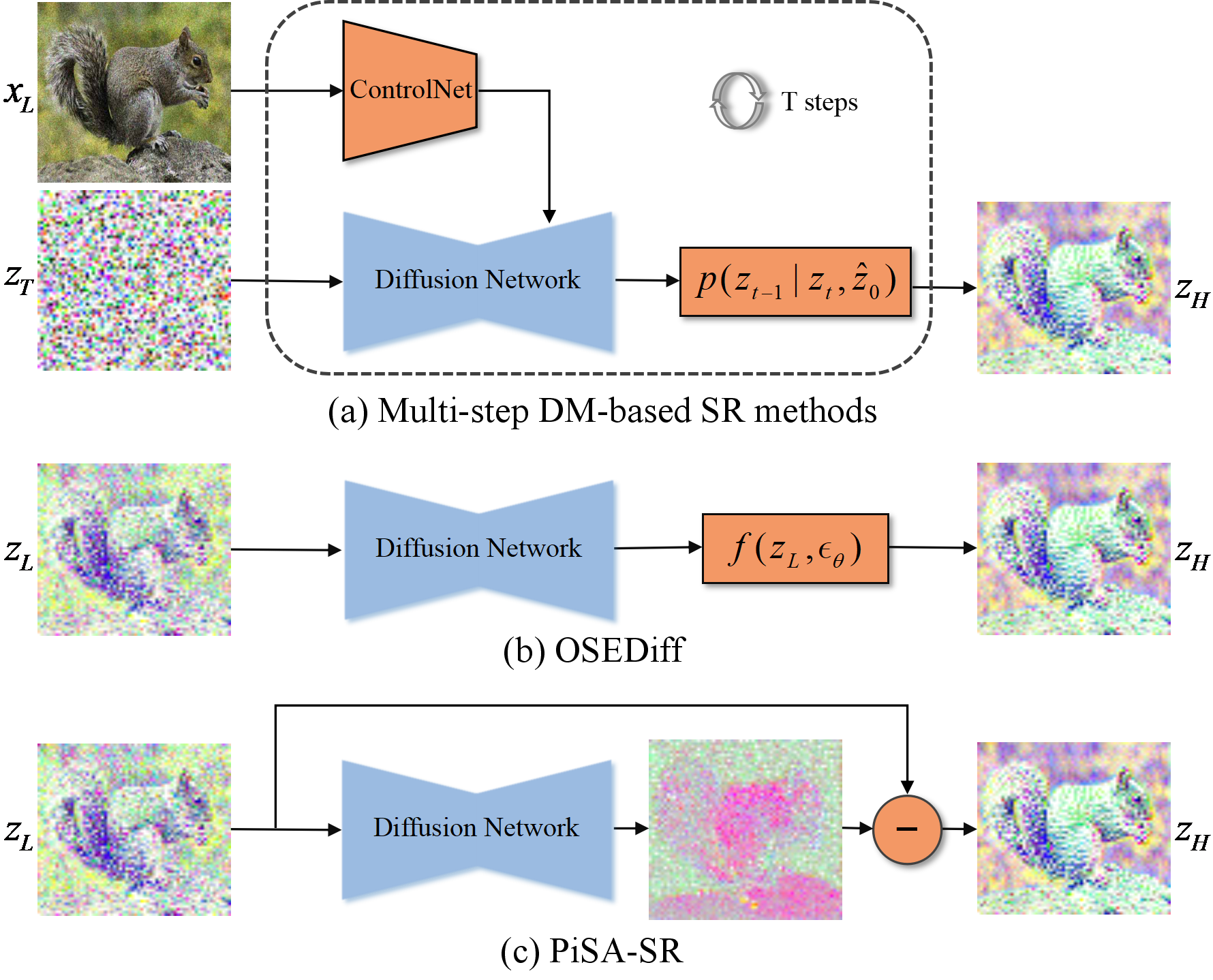}
     \vspace{-1em}
    	\caption{Comparison of the pipeline of different DM-based SR methods. (a) Multi-step methods \cite{stablesr,diffbir,pasd,seesr,xpsr,dreamclear} perform $T$ denoising steps starting from Gaussian noise $z_T$, conditioned on the LQ image $x_L$. (b) OSEDiff \cite{osediff} starts from LQ latent representation $z_L$ with only one-step diffusion. (c) Our proposed PiSA-SR formulates the SD-based SR as learning the residual between the LQ latent $z_L$ and HQ latent $z_H$.}
    	\label{fig:onestep-form}
 \vspace{-1.em}
\end{figure}

\vspace{-0.3em}
DMs such SD \cite{stablediffusion} apply a $T$-step forward process to progressively transform an initial latent code $z_0$ into Gaussian noise $z_T$ over $T$ steps. In each timestep $t$, the noisy latent code $z_t$ is obtained by directly adding noise $\epsilon \sim \mathcal{N}(0, I)$ to $z_0$ by ${z_t} = \sqrt {{{\bar \alpha }_t}}  \cdot {z_0} + \sqrt {1 - {{\bar \alpha }_t}}  \cdot \epsilon $, where $\bar {\alpha }_t$ is the cumulative parameter controlling the noise level at timestep $t$. During the reverse process, multi-step SD-based SR models \cite{stablesr,diffbir,seesr,pasd,xpsr,dreamclear} usually perform a $T$-step denoising process to gradually transform Gaussian noise $z_T$ into the HQ latent code $z_H$, conditioned on the LQ image $x_L$ with a ControlNet \cite{controlnet} (see Fig. \ref{fig:onestep-form} (a)). At stage $t$, the latent $\hat z_0$ can be computed from the noise estimated by SD UNet $\epsilon_\theta(z_t, x_L,t)$ and the current latent $z_t$ as follows:
\begin{equation}
{{\hat z}_0} = f(z_t,\epsilon_\theta)=\frac{{{z_t} - \sqrt {1 - {{\bar \alpha }_t}} \epsilon_\theta(z_t, x_L,t) }}{{\sqrt {{{\bar \alpha }_t}} }}.
\label{equ:z0}
\end{equation}
Then the latent $z_{t-1}$  at timestep $t-1$ can be sampled from $p\left( {{z_{t - 1}}\left| {{z_t},{\hat z_0}} \right.} \right) = N\left( {{z_{t - 1}};\mu_t\left( {{z_t},{\hat z_0}} \right),\sigma _t^2} \right)$, where $\mu_t$ and $\sigma _t^2$ denote the mean and variance of $z_{t-1}$, respectively. 

However, these multi-step DM-based SR methods are computationally expensive and often yield unstable results due to the random noise sampled in the diffusion process \cite{seesr,ccsr}.  To address this issue, OSEDiff  \cite{osediff} starts from the LQ latent code $z_L$ and completes the SR process with only one diffusion step (see Fig. \ref{fig:onestep-form} (b)), improving both the efficiency and stability. The transformation between $z_{L}$ and $z_{H}$ can be described as follows: 
\begin{equation}
{z_{H}} = f(z_L,\epsilon_\theta) = \frac{{{z_{L}} - \sqrt {1 - {{\bar \alpha }_{t}}} \epsilon_\theta(z_{L}) }}{{\sqrt {{{\bar \alpha }_{t}}} }}.
\label{equ:osediff}
\end{equation}
Note that we omit the timestep $t$ in the above formula since $t$ can be viewed as a constant in one-step diffusion. 

Learning the residual between the LQ and GT features has been successfully applied in deep learning-based restoration methods \cite{dncnn, zhang2018residual}. However, DMs typically perform multi-step iterations, making applying residual learning difficult. The recently developed OSEDiff \cite{osediff} employs a single diffusion step to achieve end-to-end SR training, making it feasible to adopt the residual learning strategy.
Here, we formulate the SR problem as learning the residual between $z_L$ and $z_H$, as illustrated in Fig. \ref{fig:onestep-form} (c). 
We use subtraction in the global residual connection because DMs are trained to denoise in the inverse diffusion process.
Such a residual learning formulation helps the model focus on learning the desired high-frequency information from the HQ latent features, avoiding extracting less relevant information from the LQ latent. It can also accelerate the convergence of the model training process \cite{resnet}. Furthermore, we introduce a scaling factor $\lambda$ to adjust the residual $\epsilon_\theta$ added to the LQ latent, which can be expressed as:
\begin{equation}
z_{H}=z_{L} - \lambda\epsilon_\theta(z_{L}).
\label{equ:pisa}
\end{equation}
The $\lambda$ is fixed as $1$ during training. Users can adjust the output in the inference stage by using a smaller $\lambda$ (to preserve more original contents) or a higher $\lambda$ (to enhance details more aggressively) based on their preferences.

\subsection{Dual-LoRA Training}

Previous SD-based SR methods mostly entangle the pixel-level and semantic-level enhancement in the training process \cite{stablesr,osediff}, making the balance of content fidelity and perpetual quality difficult. 
Some works sequentially perform pixel-level and semantic-level enhancement using two separate networks \cite{diffbir}. However, errors in the earlier stage may propagate to the next stage, limiting overall performance.
Furthermore, using two separate networks increases the computational burden and memory requirements, making the approach less efficient. 
It has been successfully applied in customized T2I tasks using distinct LoRA modules for various generation styles \cite{shah2025ziplora,jones2024customizing, stracke2024ctrloralter}.
Inspired by this, we propose a decoupled training approach by utilizing two LoRA modules under the pre-trained SD model, targeting pixel-level and semantic-level enhancement for the SR task respectively. Our approach, namely Dual-LoRA, introduces only a small number of additional parameters during training. These LoRA parameters can also be merged into the pre-trained SD model during inference.

\begin{figure*}
	\centering 
    	\includegraphics[scale=0.6]{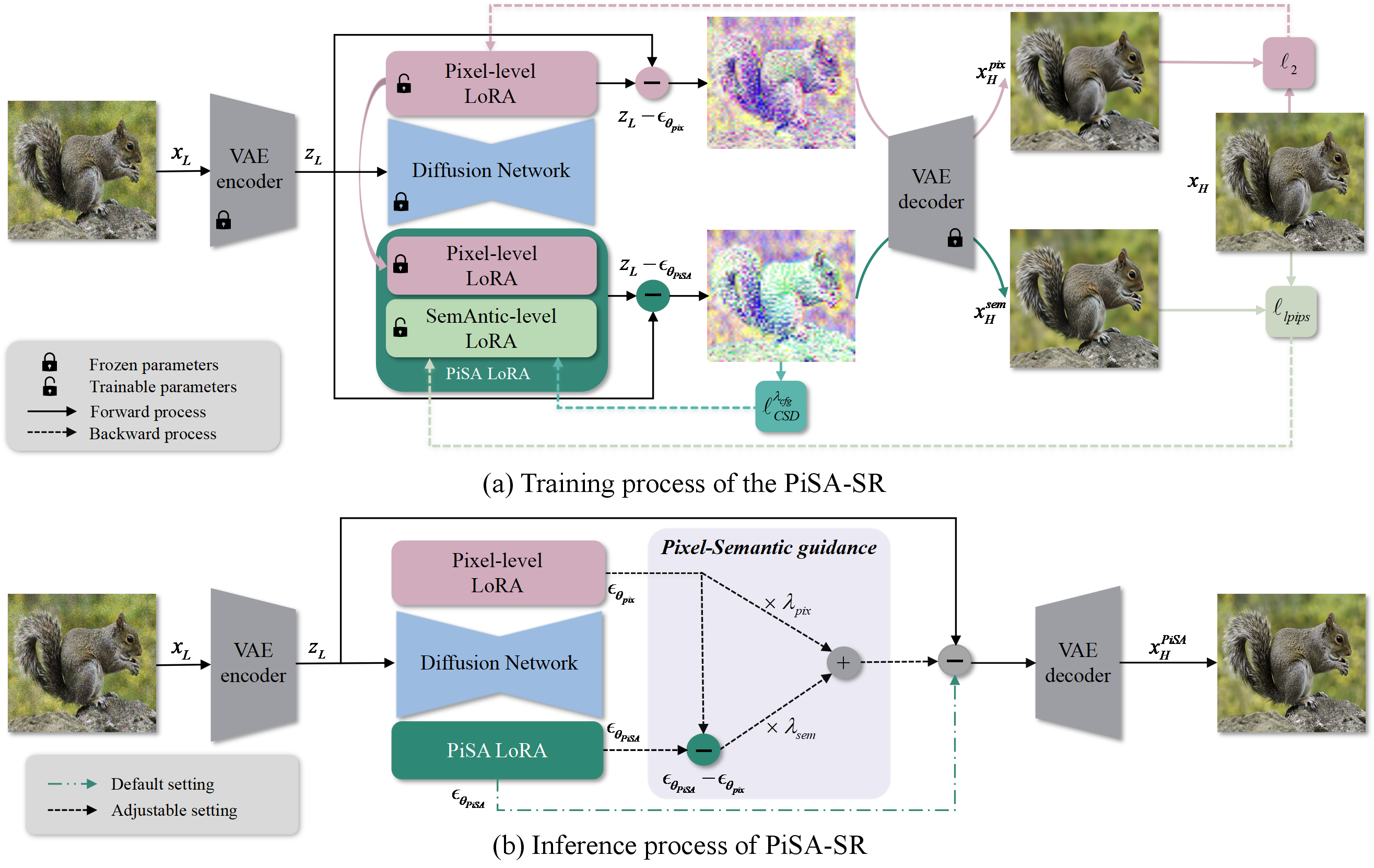}
        \vspace{-1.em}
    	\caption{The (a) training and (b) inference procedures of PiSA-SR. During the training process, two LoRA modules are respectively optimized for pixel-level and semantic-level enhancement. During the inference stage, users can use the default setting to reconstruct the HQ image in one-step diffusion or adjust $\lambda_{pix}$ and $\lambda_{sem}$ to control the strengths of pixel-level and semantic-level enhancement. }
    	\label{fig:framework}

\end{figure*}

Fig. \ref{fig:framework} (a) illustrates the training process of our approach. We freeze the parameters of the pre-trained VAE and introduce two trainable LoRA modules to the UNet of SD. Since the LQ image is corrupted by degradations such as noise, blur, and downsampling, we optimize the pixel-level LoRA first to reduce the effect of corruptions, followed by the optimization of semantic-level LoRA. 
We train the pixel-level LoRA, denoted by $\Delta \theta_{pix}$, by the pixel-level loss function. Together with the pre-trained SD parameters, the full parameter set can be represented as $ \theta_{pix} = \{\theta_{sd}, \Delta \theta_{pix}\}$. The HQ latent can then be estimated by $z_H^{pix}=z_L-\epsilon_{\theta_{pix}}(z_L)$, and decoded by the VAE decoder with $x_{H}^{pix}=D(z_H^{pix})$. 

For semantic-level enhancement, we train another  LoRA, denoted by $\Delta \theta_{sem}$, by the semantic-level loss function. 
To disentangle the pixel-level and semantic-level objectives, we fix the already trained pixel-level LoRA $\Delta \theta_{pix}$, and combine it with the semantic-level LoRA $\Delta \theta_{sem}$ to be trained, forming a PiSA-LoRA group (see Fig. \ref{fig:framework} (a)). Note that only the semantic-level LoRA module within the PiSA-LoRA group is updated in this stage. This ensures that the optimization process focuses on the semantic details without much interference from the pixel-level degradations. The full parameter set during this PiSA-LoRA training stage can be expressed as $ \theta_{P\!i\!S\!A} = \{\theta_{sd}, \Delta \theta_{pix}, \Delta \theta_{sem}\}$. The HQ latent after this stage is given by $z_H^{sem}=z_L-\epsilon_{\theta_{P\!i\!S\!A}}(z_L)$, which is then decoded by the VAE decoder with $x^{sem}_{H}=D(z_H^{sem})$.

\begin{figure}
	\centering 
    	\includegraphics[scale=0.82]{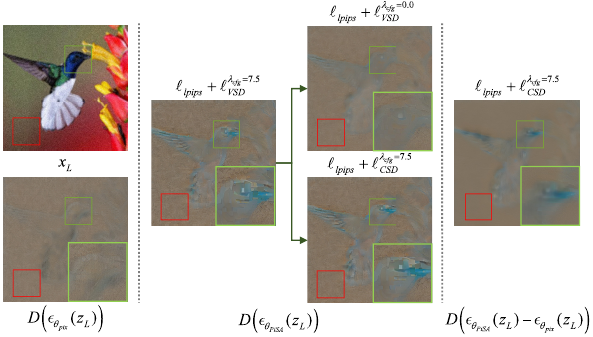}
     \vspace{-1.em}
    	\caption{The model outputs with pixel-wise and semantic-level losses for a given LQ image. 
     }
    	\label{fig:res}
 \vspace{-0.5em}
\end{figure}

\subsection{Pixel-level and Semantic-level Losses}

We train the pixel-level LoRA using the $\ell_2$ loss. As shown by $D(\epsilon_{\theta_{pix}}(z_L))$ in Fig. \ref{fig:res}, the $\ell_2$ loss can effectively remove degradation and enhance edges. However, it is insufficient to generate semantic-level details \cite{srgan,esrgan,LDL}, resulting in smooth SR outputs. 
The LPIPS loss \cite{lpips} can be used to activate semantic details by aligning the high-level features with a pre-trained classification VGG network \cite{vgg}, which is however trained on a limited set of image categories. The GAN loss \cite{srgan} captures semantic information through adversarial training, encouraging the generator to produce more realistic images. However, GAN loss can be unstable in training and generate much artifacts \cite{LDL}.

Recently developed SD models \cite{stablediffusion} have shown superior performance in generating open-category images with enhanced semantic details. As a conditional generation model, SD can be used to form up an implicit classifier \cite{cfg} for the posterior distribution modeling, and its gradient w.r.t. the synthesized image ${{x}}_t$ is as follows:

\begin{equation}
    \begin{aligned}
        \nabla_{x_t} \log p_\theta(c \mid x_t) 
        &= \nabla_{x_t} \log p_\theta(x_t \mid c) - 
            \nabla_{x_t} \log p_\theta(x_t) \\
        &= - (\epsilon_\theta(x_t, c, t) - \epsilon_\theta(x_t, t)) / \sigma_t,
    \end{aligned}
    \label{eq:SD_class}
\end{equation}
where $c$ is the text prompt,  
$t$ is the current timestep, and $\sigma_t=\sqrt {1 - {{\bar \alpha }_t}}$.

The SD model, parameterized by $\theta$, model score functions of $\nabla_{x_t} \log p_\theta(x_t \mid c)$ and  $\nabla_{x_t} \log p_\theta(x_t)$ using noise predictions~\cite{song2020score}.
By taking the expectation of the gradient in \cref{eq:SD_class} over all possible $t$, we have the CSD~\cite{cfg} loss, which is originally used in 3D generation tasks to optimize the posterior probability of rendered images aligning their semantic content with text prompts.

Motivated by the functionality of CSD loss in DMs for generation, we investigate the usage of CSD loss for semantic-level enhancement in the SR task. Following \cite{dmd, osediff}, and for the clarity of the following context, we formulate the gradient of CSD in the latent space instead of the noise domain. Specifically, the gradient of CSD loss can be formulated in Eq. (\ref{equ:csd}) below (the detailed derivation can be found in the \textbf{supplementary materials}): 
\begin{equation}
\footnotesize{
\nabla \ell _{CSD}^{{\lambda _{c\!f\!g}}} = {\mathbb{E}_{t,\epsilon,{z_t},c}}\left[ {w_t \left(f({z_t},\epsilon_{real})-f({z_t},\epsilon_{real}^{\lambda_{c\!f\!g}}) \right)\frac{{\partial {z_H^{sem}}}}{{\partial {\theta _{P\!i\!S\!A}}}}} \right]\!},
\label{equ:csd}
\end{equation}
where the gradient expectation is calculated considering all diffusion timesteps  $t={1,...,T}$, the noise is sampled from $\epsilon \sim \mathcal{N}(0, I)$, $z_t$ is obtained by ${z_t} = \sqrt {{{\bar \alpha }_t}}  \cdot {z_H^{sem}} + \sqrt {1 - {{\bar \alpha }_t}}  \cdot \epsilon $, $c$ is the text prompt extracted from $z_H^{sem}$, $f(\cdot)$ is the function as Eq. (\ref{equ:z0}), $w_t$ is the timestep-dependent scalar weight, and $\epsilon_{real}^{{\lambda _{c\!f\!g}}}$ denotes the pre-trained SD output with the CFG term $\epsilon_{real}^{cls}$ same as Eq. (\ref{eq:SD_class}) \cite{cfg,csd}:
\begin{equation}
\footnotesize{
    \epsilon_{real}^{{\lambda _{c\!f\!g}}} = {\epsilon_{real}}\left( {{z_t},t} \right) + \underbrace {{\lambda _{c\!f\!g}}\left( {{\epsilon_{real}}\left( {{z_t},t,c} \right) - {\epsilon_{real}}\left( {{z_t},t} \right)} \right)}_{\epsilon_{real}^{cls}\left( {{z_t},t,c} \right)}\!.}
\label{equ:cfgterm}
\end{equation}

Note that the VSD loss \cite{vsd, dmd}, which also aims to align the distributions of enhanced and natural images within the latent space, has been verified effective for SR by OSEDiff \cite{osediff}. The gradient of VSD loss can be written as:
\begin{equation}
\footnotesize{
\nabla \ell_{V\!S\!D}^{{\lambda _{c\!f\!g}}} = {\mathbb{E}_{t,\epsilon,{z_t},c}}\left[ { w_t \left( {f({z_t},{\epsilon_{fake}})-f({z_t},\epsilon_{real}^{{\lambda _{c\!f\!g}}})} \right)\frac{{\partial {z_H^{sem}}}}{{\partial {\theta _{P\!i\!S\!A}}}}} \right]\!,}
\label{equ:vsd}
\end{equation}
where $\epsilon_{fake}$ denotes the output of fine-tuned SD aligned with the distribution of synthetic images. 
The VSD loss with $\lambda_{c\!f\!g}$ can be divided into two components: the VSD loss with $\lambda_{c\!f\!g}=0.0$, \ie, $\ell_{VSD}^{\lambda_{cfg}=0.0}$, and the CSD loss. 
Coupling with the LPIPS loss, we visualize the semantic-level LoRA optimization results of the two components of VSD in Fig. \ref{fig:res}. We see that the CSD loss with the normal guidance scale of $\lambda_{c\!f\!g}=7.5$ contributes more significantly to semantic enhancement. In contrast, the VSD loss with $\lambda_{c\!f\!g}=0.0$ weakens the semantic details.
More importantly, the component $\ell_{VSD}^{\lambda_{cfg}=0.0}$ requires bi-level optimization, leading to intensive memory usage and unstable training \cite{ma2025scaledreamer}. In comparison, the CSD loss is free of bi-level optimization, reducing significantly memory usage and improving training stability. 
Therefore, we integrate LPIPS and CSD losses for the semantic-level LoRA optimization.

\subsection{The Inference Process of PiSA-SR}
\vspace{-0.3em}

Fig. \ref{fig:framework}(b) shows the inference process of PiSA-SR. In the default inference setting (represented by the green dash-dot line), only the PiSA-LoRA, which merges the pixel-level and semantic-level LoRA modules, is used to process the input together with the pre-trained SD model, achieving state-of-the-art SR performance in just one-step diffusion (see our experiments in Section \ref{subsec:comparison}). 
To achieve flexible SR with diverse user preferences, we extend the strategy in Eq. (\ref{equ:pisa}) by introducing a pair of pixel and semantic guidance factors, denoted by $\lambda_{pix}$ and $\lambda_{sem}$, to control the SR results:
\begin{equation}
\epsilon_\theta(z_{L}) =
\lambda_{pix} \epsilon_{\theta_{pix}}(z_{L}) + \lambda_{sem} (\epsilon_{\theta_{P\!i\!S\!A}}(z_{L}) - \epsilon_{\theta_{pix}}(z_{L})). 
\label{equ:ps_guidance}
\end{equation}

In Eq. (\ref{equ:ps_guidance}), $\epsilon_{\theta_{pix}}(z_L)$ is the output with only pixel-level LoRA, while $\epsilon_{\theta_{P\!i\!S\!A}}(z_L)$ is the output with both pixel and semantic level enhancement. Inspired by CFG \cite{cfg}, the difference between them, \ie, $\epsilon_{\theta_{P\!i\!S\!A}}(z_L)-\epsilon_{\theta_{pix}}(z_L)$, can well separate pixel-level degradations from semantic details (see the red box in Fig. \ref{fig:res}). Therefore, as illustrated in Fig. \ref{fig:framework}(b) (represented by the black dashed line), by adjusting $\lambda_{pix}$ and $\lambda_{sem}$, we can control the contributions of pixel-level LoRA and semantic-level LoRA, and consequently deliver the SR results with different styles, as shown in Fig. \ref{fig1}.

%% file: sec/4_Experiement.tex
\section{Experiments}
\label{sec:Experiment}
\vspace{-0.3em}
\subsection{Experimental Settings}
\vspace{-0.3em}
\noindent\textbf{Training settings.}
We train PiSA-SR upon SD 2.1-base for the $\times4$ SR task. The two trainable LoRA modules are applied to the weights of all convolution and MLP layers. Both modules are initialized using a Gaussian distribution with a rank of 4. Following SeeSR \cite{seesr} and OSEDiff \cite{osediff}, we employ LSDIR \cite{lsdir} and the first 10K images from FFHQ \cite{ffhq} as the training data. We generate paired training data using the degradation pipeline from RealESRGAN \cite{realesrgan}. The batch size is 16 and the training patch size is 512×512. The Adam optimizer \cite{adam} is used with a learning rate 5e-5. The pixel-level and semantic-level LoRA modules undergo 4K and 8.5K training iterations. 

\noindent\textbf{Compared methods.} 
We compare PiSA-SR with leading multi-step DM-based methods StableSR \cite{stablesr}, ResShift \cite{resshift}, DiffBIR \cite{diffbir}, PASD \cite{pasd}, and SeeSR \cite{seesr}, one-step DM-based methods SinSR \cite{sinsr} and OSEDiff \cite{osediff}, and GAN-based SR methods RealESRGAN \cite{realesrgan}, BSRGAN \cite{bsrgan} and LDL \cite{LDL}. All comparative results are obtained using officially released codes or models. Due to space constraints, comparisons with GAN-based methods are presented in the \textbf{supplementary materials.}

\noindent\textbf{Test datasets and evaluation metrics.} 
Following previous arts \cite{stablesr, seesr, osediff}, we evaluate the competing methods using synthetic and real-world test data. The synthetic dataset comprises 3000 images cropped from DIV2K \cite{div2k} with size $512 \times 512$, degraded using RealESRGAN degradation pipeline \cite{realesrgan}. The real-world data are center-cropped from RealSR \cite{realsr} and DrealSR \cite{drealsr} datasets with size $128 \times 128$ for LQ images and $512 \times 512$ for HQ images. 
PSNR and SSIM \cite{ssim}, computed on the Y channel in the YCbCr space, are used to measure the fidelity of SR results; LPIPS \cite{lpips} and DISTS \cite{dists}, computed in the RGB space, are used to assess the perceptual quality of SR results; FID \cite{fid} evaluates the distance of distributions between GT and restored images; NIQE \cite{niqe}, CLIPIQA \cite{clipiqa}, MUSIQ \cite{musiq} and MANIQA \cite{maniqa} evaluate the image quality without reference images. Additionally, a user study is conducted to validate the effectiveness of the proposed adjustable SR in the \textbf{supplementary materials}.

\input{table/adjustable_comparison}

\input{table/comparison}

\subsection{Experiments on Adjustable SR}
\vspace{-0.3em}
We conduct experiments by fixing one guidance scale ($\lambda_{pix}$ or $\lambda_{sem}$) at 1 and varying the other to observe how the reconstructed images change. 
PSNR, LPIPS, CLIPIQA, and MUSIQ are used to evaluate performance. PSNR measures pixel-level fidelity; LPIPS evaluates the image perceptual quality with GT as reference; CLIPIQA and MUSIQ measure the image quality without using a reference. The results are shown in Table \ref{tab:adjustable}.
First, we observe that increasing the pixel-level scale $\lambda_{pix}$ leads to a continuous improvement in the no-reference metrics CLIPIQA and MUSIQ. 
This is because increasing $\lambda_{pix}$ can remove image degradation and enhance edge, which is favorable to these two metrics. 
However, the reference-based metrics PSNR and LPIPS exhibit a rise-and-fall pattern. PSNR peaks at $\lambda_{pix} = 0.5$, indicating the best pixel-level fidelity. LPIPS reaches its best value at $\lambda_{pix} = 0.8$ (0.2612), suggesting that at this point the restored image is perceptually most similar to the GT with rich details. Further increasing $\lambda_{pix}$ reduces both PSNR and LPIPS scores.
Second, increasing the semantic-level scale $\lambda_{sem}$ also results in a continuous improvement in CLIPIQA and MUSIQ, but with a higher upper bound than pixel-level adjustments. This is because increasing $\lambda_{sem}$ can synthesize more semantic-level details (see Fig. \ref{fig1}). However, PSNR decreases as $\lambda_{sem}$ increases, while LPIPS first improves and peaks at $\lambda_{sem} = 0.8$ (0.2465), and then declines. This is because excessive semantic details may induce changes in image content and reduce pixel-level fidelity. Meanwhile, the over-enhanced images may exhibit many details different from the GT, making the reference-based LPIPS index deteriorate.

Fig. \ref{fig1} illustrates the visual comparisons of different pixel-semantic scales. More comparisons can be found in the \textbf{supplementary materials.}
From bottom to top, as the pixel-level scale $\lambda_{pix}$ increases, the model progressively removes degradation from the input LQ image, resulting in clearer and sharper visual output. However, a too-high $\lambda_{pix}$ can lead to loss of details, and diminish the effectiveness of semantic improvements, as seen in the top-right corner of Fig. \ref{fig1}.
From left to right, increasing the semantic-level scale $\lambda_{sem}$ enhances the semantic richness of the image, making the details of the wrinkles, beard, and hair of Albert Einstein more pronounced.
However, an excessive semantic level such as $\lambda_{sem}$=1.5 introduces unnatural artifacts, making the SR result less realistic. 
With the flexibility of PiSA-SR, users can customize their preferences, either preserving more fidelity or emphasizing richer semantic enhancements, depending on their specific use case.

\begin{figure*}
	\centering 
    	\includegraphics[scale=1.1]{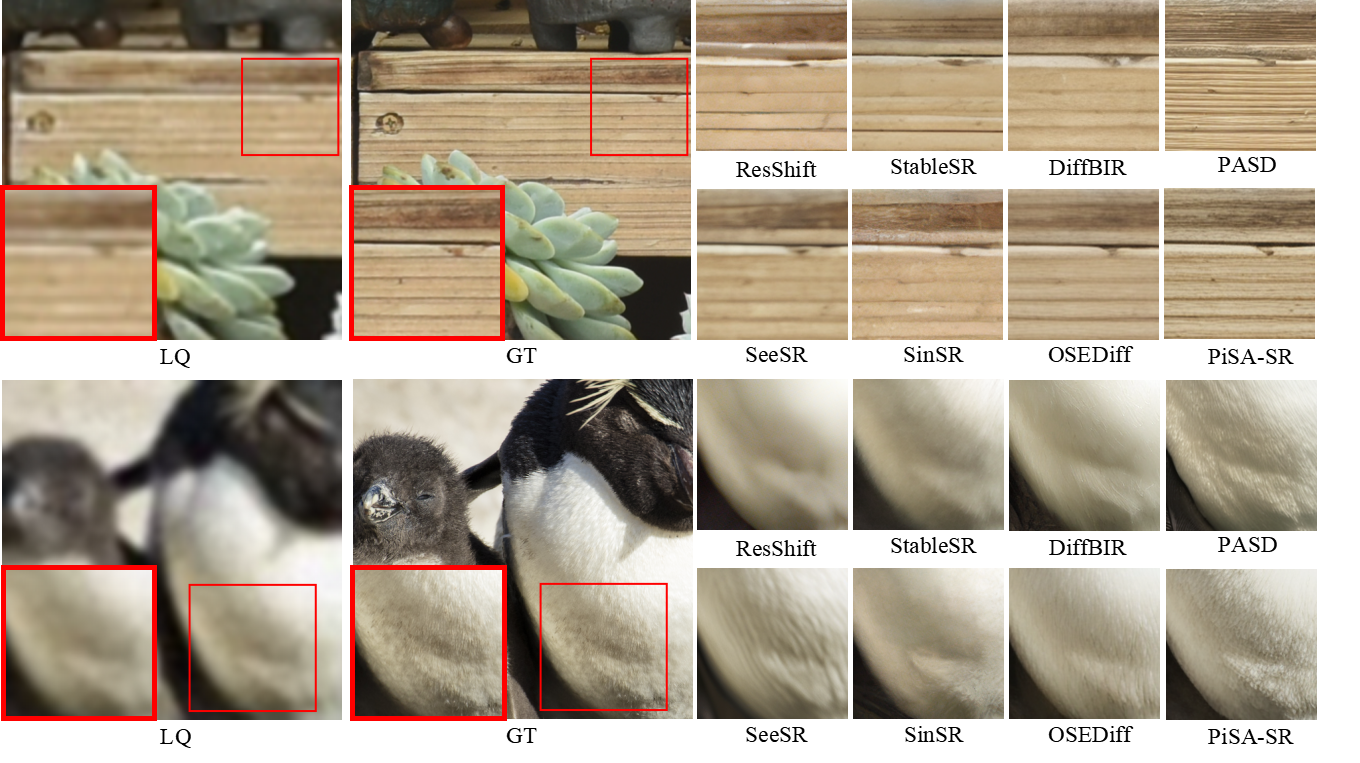}
     \vspace{-1em}
    	\caption{Visual comparisons of different DM-based SR methods. Please zoom in for a better view.}
    	\label{fig:comparison}
 \vspace{-1em}
\end{figure*}

\input{table/inference_time}

\subsection{Comparisons with State-of-the-Arts}
\label{subsec:comparison}
\vspace{-0.3em}

As shown in Eq. (\ref{equ:ps_guidance}), letting $\lambda_{pix}=1$ and $\lambda_{sem}=1$, we have the default version of PiSA-SR, which can be performed in just one-step diffusion. Here, we compare this default version of PiSA-SR with the competing methods. 

\noindent\textbf{Quantitative comparisons.}
Table \ref{tab:comparison} compares the performance of our default PiSA-SR model with other DM-based Real-SR methods. The following findings can be made. ResShift \cite{resshift} and its distilled version SinSR \cite{sinsr} show limited performance in no-reference metrics, implying poor perceptual quality. 
PASD \cite{pasd} and SeeSR \cite{seesr} leverage the pre-trained SD as their base model, and extract additional high-level information \cite{RAM} to improve the perceptual quality of images. Their no-reference metric scores, such as MUSIQ and CLIPIQA, are highly improved. However, the multi-step inference makes them less efficient. In addition, their relatively low LPIPS and DISTS scores suggest their less faithful reconstruction to the GT. 
OSEDiff \cite{osediff} distills the generation capacity of multi-step DM into just one-step diffusion by the VSD loss \cite{vsd, dmd}. While effective and efficient, its no-reference metrics are limited. 
Our proposed PiSA-SR also exhibits high efficiency by requiring only one inference step, while showing impressive pixel-level fidelity and semantic-level perceptual quality. It outperforms other methods not only in reference-based metrics such as LPIPS and DISTS but also in no-reference metrics such as CLIPIQA, MUSIQ, and MANIQA.

\noindent\textbf{Qualitative comparisons.} 
We present visual comparisons in Fig. \ref{fig:comparison}. ResShift and SinSR fail to generate richer textures and details because of their limited generation capacity. StableSR and DiffBIR can generate finer details (\eg, more realistic penguin feathers in the second example), due to the leveraged strong SD prior. PASD and SeeSR incorporate high-level information guidance, resulting in more comprehensive details. However, the inherent randomness in PASD and SeeSR may lead to over-enhanced (\eg, unnatural penguin feathers generated by SeeSR) or over-smoothed details (\eg, the blurred texture of the wooden stool generated by SeeSR).
OSEDiff can generate more consistent results in both examples. However, the restored details can be limited.
In contrast, the proposed PiSA-SR delivers high-quality, realistic SR results. With the dual-LoRA design, the restored structures align well with the input LQ image, providing more accurate fidelity information and generating more natural and richer details. More visualization comparisons can be found in \textbf{supplementary materials}.

\noindent\textbf{Complexity comparisons}.
We compare the number of parameters and inference time of competing DM-based SR models in Table \ref{tab:inferencetime}, where PiSA-SR-def. and PiSA-SR-adj. represent its default setting and adjustable setting, respectively. Inference time is measured on the $\times 4$ SR task with $128 \times 128$ LQ images using a single NVIDIA A100 80G GPU.
ResShift has fewer parameters because it is trained from scratch without using the pre-trained SD model. SinSR inherits the parameters from its parental ResShift model. Among the SD-based SR methods, PiSA-SR-def. has the fewest parameters and achieves the fastest inference time. Unlike OSEDiff, PiSA-SR-def. does not require a semantic extractor (\eg, RAM \cite{RAM}) during inference. Instead, it distills the semantic information into the semantic-level LoRA with CSD. 
PiSA-SR-adj., which requires two diffusion steps to compute outputs from pixel-level and semantic-level LoRA modules, incurs a slightly longer inference time over the default setting. However, this small increase in time offers users the advantage of generating SR outputs tailored to their specific needs.

\noindent\textbf{Ablation study}.
We performed a series of ablation studies for a better understanding of the dual-LoRA training approach, and the roles of the two LoRA modules. The details can be found in the \textbf{supplementary materials}. 

%% file: table/adjustable_comparison.tex
\begin{table}\tiny
\caption{Results of PiSA-SR with different pixel-semantic guidance scales on the RealSR test dataset.}
  \vspace{-1.5em}
\centering
 
\resizebox{\linewidth}{!}{
\begin{tabular}{cc|cccc}

 \hline
\cellcolor{lightpurple}$\lambda_{pix}$   & \cellcolor{lightgreen}$\lambda_{sem}$     & PSNR↑ & LPIPS↓  & CLIPIQA↑ & MUSIQ↑ \\ \hline
\cellcolor{lightpurple}0.0 & 1.0 &   25.96   &   0.3426 &  0.4129 &  46.45 \\
\cellcolor{lightpurple}0.2 & 1.0 &   26.48  &  0.3042 &  0.4868 & 54.05   \\
\cellcolor{lightpurple}0.5 & 1.0 &   26.75   &   0.2646  &  0.5705 &   63.82   \\
\cellcolor{lightpurple}0.8 & 1.0 &   26.18   &   0.2612  &  0.6292 &   68.95    \\
\cellcolor{lightpurple}1.0 & 1.0 &   25.50   &   0.2672    &   0.6702 &  70.15 \\ 
\cellcolor{lightpurple}1.2 & 1.0 &   24.76   & 0.2723  &  0.6746  &   70.33   \\ 
\cellcolor{lightpurple}1.5 & 1.0 &   23.74  & 0.2769  &  0.6305  &   69.23   \\ \hline
1.0 & \cellcolor{lightgreen}0.0 &   26.92   &  0.3018 & 0.3227 & 49.62   \\
1.0 & \cellcolor{lightgreen}0.2 &   26.95  &  0.2784 &  0.3591 & 53.64  \\
1.0 & \cellcolor{lightgreen}0.5 &   26.77   &  0.2476 &  0.4322 & 58.76   \\
1.0 & \cellcolor{lightgreen}0.8 &   26.20   &    0.2465 &  0.5806 & 66.33   \\
1.0 & \cellcolor{lightgreen}1.0 &   25.50   &   0.2672    &   0.6702 &  70.15  \\
1.0 & \cellcolor{lightgreen}1.2 & 24.59 &   0.3000   &   0.7015 &   71.60  \\ 
1.0 & \cellcolor{lightgreen}1.5 & 23.08 &   0.3541  &  0.6835 &   71.76     \\ \hline
\end{tabular}
}
\label{tab:adjustable}
 \vspace{-2em}
\end{table}

%% file: table/comparison.tex
\begin{table*}\tiny
           \centering

 \caption{
        Quantitative comparison among the state-of-the-art DM-based SR methods on synthetic and real-world test datasets. `S' denotes the number of diffusion steps. The best and the second-best results are highlighted in {\color[HTML]{FF0000} \textbf{red}} and  {\color[HTML]{2E75B5} \textbf{blue}}, respectively.}
  \vspace{-1.em}
   \resizebox{\linewidth}{!}{
\begin{tabular}{c|c|ccccccccc}
\hline
Datasets                   & Methods                                                  & PSNR↑&SSIM↑& LPIPS↓ & DISTS↓ & FID↓    & NIQE↓ & CLIPIQA↑ & MUSIQ↑                                  & MANIQA↑                                                   \\

\hline
                           & ResShift-S15                                             & {\color[HTML]{FF0000} \textbf{24.69}}          & {\color[HTML]{FF0000} \textbf{0.6175}}& 0.3374                                       & 0.2215                                 & 36.01                                  & 6.82                                          & 0.6089                                       & 60.92                                &                                               0.5450                     \\
                           
                           & StableSR-S200                                            & 23.31                                      & 0.5728                              & 0.3129                                     & 0.2138                                  & {\color[HTML]{FF0000} \textbf{24.67}}  & 4.76                                       & 0.6682                                        & 65.63                                 &                                               0.6188                       \\
                           
                           & DiffBIR-S50                                              & 23.67                 & 0.5653                                        & 0.3541                                         & 0.2129                                       & 30.93                                  & 4.71& 0.6652                                     & 65.66                            &                                               0.6204                              \\
                           
                           & PASD-S20                                                 & 23.14               & 0.5489                                        & 0.3607                                       & 0.2219                                      & 29.32                                  & {\color[HTML]{FF0000} \textbf{4.40}}       & 0.6711                                         & {\color[HTML]{0070C0} \textbf{68.83}}&                                               {\color[HTML]{FF0000} \textbf{0.6484}}                             \\
                           
                           & SeeSR-S50                                                & 23.71                                     & 0.6045                             & 0.3207                                  & {\color[HTML]{0070C0} \textbf{0.1967}}& 25.83 & 4.82                              & {\color[HTML]{0070C0} \textbf{0.6857}}& 68.49                                 & 0.6239                                \\
                           
& SinSR-S1                                                 & {\color[HTML]{0070C0} \textbf{24.43}}& 0.6012                                 & 0.3262                             & 0.2066                                         & 35.45                                  & 6.02                                      & 0.6499                        & 62.80                               & 0.5395                                \\

 & OSEDiff-S1                                               & 23.72                                                & {\color[HTML]{0070C0} \textbf{0.6108}}& {\color[HTML]{0070C0} \textbf{0.2941}}& 0.1976& 26.32&      4.71         & 0.6683                                            & 67.97                                      & 0.6148                   \\
\rowcolor{lightpink}
 \multirow{-8}{*}{DIV2K}& PiSA-SR-S1& 23.87& 0.6058& {\color[HTML]{FF0000} \textbf{0.2823}}& {\color[HTML]{FF0000} \textbf{0.1934}}& {\color[HTML]{0070C0} \textbf{25.07}}& {\color[HTML]{0070C0} \textbf{4.55}}& {\color[HTML]{FF0000} \textbf{0.6927}}& {\color[HTML]{FF0000} \textbf{69.68}}& {\color[HTML]{0070C0} \textbf{0.6400}}\\
\hline 
                           & ResShift-S15                                             &     {\color[HTML]{FF0000} \textbf{26.31}}& {\color[HTML]{0070C0} \textbf{0.7411}}& 0.3489                             & 0.2498                                         & 142.81                                 & 7.27                                       & 0.5450                                & 58.10                           &                                               0.5305                                \\
                           & StableSR-S200                                            & 24.69                                      & 0.7052                 & 0.3091                     & 0.2167                                     & 127.20                                 & 5.76                                    & 0.6195                   & 65.42       &                                               0.6211                      \\
                           & DiffBIR-S50                                              & 24.88       & 0.6673                         & 0.3567                   & 0.2290                            & 124.56                                 & 5.63                                 & 0.6412                          & 64.66                            &                                               0.6231                \\
                           & PASD-S20                                                 & 25.22              & 0.6809                   & 0.3392                      & 0.2259                             & {\color[HTML]{FF0000} \textbf{123.08}}& {\color[HTML]{FF0000} \textbf{5.18}}          & 0.6502                                 & 68.74                         &                                               {\color[HTML]{0070C0} \textbf{0.6461}}   \\
 & SeeSR-S50                                                & 25.33& 0.7273                                         & 0.2985                                         & 0.2213                                     & 125.66                                 & {\color[HTML]{0070C0} \textbf{5.38}}          &0.6594     & {\color[HTML]{0070C0} \textbf{69.37}} &0.6439\\
                           & SinSR-S1                                                 & {\color[HTML]{0070C0} \textbf{26.30}}& 0.7354& 0.3212                                    & 0.2346                                 & 137.05                                 & 6.31                                 & 0.6204                                     & 60.41                                & 0.5389             \\
                           & OSEDiff-S1                                               & 25.15                     & 0.7341                                     & {\color[HTML]{0070C0} \textbf{0.2921}}                 & {\color[HTML]{0070C0} \textbf{0.2128}}               & {\color[HTML]{0070C0} \textbf{123.50}}& 5.65& {\color[HTML]{0070C0} \textbf{0.6693}}           & 69.09&                                               0.6339               \\
\rowcolor{lightpink}
\multirow{-8}{*}{RealSR}  & PiSA-SR-S1 & 25.50& {\color[HTML]{FF0000} \textbf{0.7417}}& {\color[HTML]{FF0000} \textbf{0.2672}}& {\color[HTML]{FF0000} \textbf{0.2044}}& 124.09& 5.50& {\color[HTML]{FF0000} \textbf{0.6702}}& {\color[HTML]{FF0000} \textbf{70.15}}& {\color[HTML]{FF0000} \textbf{0.6560}}\\
  \hline 
                           & ResShift-S15                                             & {\color[HTML]{FF0000} \textbf{28.45}}                                      & 0.7632               & 0.4073              & 0.2700                                      & 175.92                                 & 8.28                                   & 0.5259                       & 49.86        &                                               0.4573               \\
                           & StableSR-S200                                            & 28.04        & 0.7460              & 0.3354                                   & 0.2287                                  & 147.03 & 6.51            & 0.6171                       & 58.50                         &                                               0.5602                                \\
                           & DiffBIR-S50                                              & 26.84             & 0.6660                     & 0.4446                 & 0.2706                                   & 167.38                                 & {\color[HTML]{0070C0} \textbf{6.02}} & 0.6292              & 60.68       &                                               0.5902                               \\
                           & PASD-S20                                                 & 27.48                        & 0.7051                            & 0.3854                            & 0.2535                                        & 157.36                                 & {\color[HTML]{FF0000} \textbf{5.57}}& 0.6714& 64.55                                 &                                               {\color[HTML]{0070C0} \textbf{0.6130}}                       \\
                           & SeeSR-S50                                                & 28.26                                      & 0.7698                                 & 0.3197                                      & 0.2306                                  & 149.86                                 & 6.52                                        & 0.6672                                  & 64.84                       & 0.6026                                        \\
                           & SinSR-S1                                                 & {\color[HTML]{0070C0} \textbf{28.41}}          & 0.7495        & 0.3741        & 0.2488                        & 177.05                                 & 7.02                                     & 0.6367                      & 55.34                          &                                               0.4898                             \\
                           & OSEDiff-S1                                               & 27.92                              & {\color[HTML]{FF0000} \textbf{0.7835}}         & {\color[HTML]{0070C0} \textbf{0.2968}}        & {\color[HTML]{FF0000} \textbf{0.2165}}               & {\color[HTML]{0070C0} \textbf{135.29}} & 6.49     & {\color[HTML]{0070C0} \textbf{0.6963}}               & {\color[HTML]{0070C0} \textbf{64.65}}                                      & 0.5899                                         
\\
\rowcolor{lightpink}
\multirow{-8}{*}{DrealSR}& PiSA-SR-S1&28.31 & {\color[HTML]{0070C0} \textbf{0.7804}}& {\color[HTML]{FF0000} \textbf{0.2960}}& {\color[HTML]{0070C0} \textbf{0.2169}}& {\color[HTML]{FF0000} \textbf{130.61}}& 6.20&{\color[HTML]{FF0000} \textbf{0.6970}} & {\color[HTML]{FF0000} \textbf{66.11}}& {\color[HTML]{FF0000} \textbf{0.6156}}\\ \hline
\end{tabular}
}
 \vspace{-2em}
\label{tab:comparison}
\end{table*}

%% file: table/inference_time.tex
\begin{table*}\tiny
\caption{
The inference time and the number of parameters of DM-based SR methods.
}
 \vspace{-1.5em}
\label{tab:inferencetime}

\centering
\resizebox{\linewidth}{!}{
\begin{tabular}{c|ccccc|cc|>{\columncolor{lightpink}}c>{\columncolor{lightpurple}}c}
\hline
                     & StableSR& ResShift  & DiffBIR & PASD & SeeSR &SinSR &OSEDiff&PiSA-SR-def.&PiSA-SR-adj. \\ \hline
Inference Steps &    200   &    15      &     50    & 20   &   50 & 1&1&1& 2\\ \hline
Inference time(s)/Image &   10.03&  0.76& 2.72   &  2.80    & 4.30  &  0.13&0.12&0.09& 0.13\\ \hline
\#Params(B) &   1.56&     0.18&   1.68&    2.31& 2.51 & 0.18&1.77&1.30&1.30\\ \hline
\end{tabular}}
 \vspace{-2em}
\end{table*}

%% file: sec/5_conclusion.tex
\section{Conclusion and Limitation}
\vspace{-0.3em}

In this work, we presented PiSA-SR, a novel SR framework that disentangles pixel-level and semantic-level objectives by learning two LoRA modules upon the pre-trained SD model. By decoupling the learning process into two distinct spaces, PiSA-SR effectively balanced pixel-wise regression and perceptual quality. The commonly used $\ell_2$ loss was used to optimize pixel-level LoRA, while the LPIPS and CSD losses were employed to optimize the semantic-level LoRA. PiSA-SR demonstrated impressive performance in both effectiveness and efficiency. Furthermore, it offered high flexibility during inference through adjustable guidance scales, allowing users to customize the SR results without model re-training.

While PiSA-SR provided a flexible solution for adjusting pixel-level fidelity and semantic-level perception, it incurs a slight inference time increment over its default version. In addition, we use a single pixel-level LoRA to handle various degradations in the LQ, which may be insufficient for heavy degradations. In the future, we will explore refining separate LoRA spaces for different image degradations and the possibility of image-adaptive scale factors.

%% file: sec/supplementary.tex

\noindent The following materials are provided in this supplementary file:
\begin{itemize}
            \item The detailed derivation from the SD classifier formulation (Eq. (4) in the main paper) to the CSD loss formulation (Eq. (5) in the main paper) (referring to Sec. 3.3 in the main paper);

            \item Comparisons with GAN-based methods (referring to Sec. 4.1 in the main paper);

            \item User study (referring to Sec. 4.1 in the main paper.)
            
            \item More visual examples of different pixel-semantic scale selections (referring to Sec. 4.2 in the main paper);
            
            \item More visual comparisons between our PiSA-SR and DM-based SR methods (referring to Sec. 4.3 in the main paper);
            
            \item Ablation studies (referring to Sec. 4.3 in the main paper).

	\end{itemize}
 
\section{The detailed derivation}
\label{sec:derivation}
We rewrite Eq. (4) in the main paper as follows:

\begin{equation}
    \begin{aligned}
       \nabla \ell _{CSD}^{{\lambda _{c\!f\!g}}}
       & = {\mathbb{E}_{t,\epsilon,{z_t},c}} \left[\frac{w_t}{{{\sigma _t}}} (\epsilon_{real}(z_t,t,c) - \epsilon_{real}(z_t,t))\frac{{\partial {z_H^{sem}}}}{{\partial {\theta _{P\!i\!S\!A}}}}\right]\!,
    \end{aligned}
    \tag{S1}
    \label{eq:SD_csd1}
\end{equation}
where the SD model is parameterized by $\theta$, $z_H^{sem}$ is the estimated HQ latent with the PiSA LoRA, $c$ is the text prompt extracted from $z_H^{sem}$, $\sigma_t=\sqrt {1 - {{\bar \alpha }_t}}$ is the standard deviation, and $\epsilon_{real}$ is the output of pre-trained SD model. 
$z_t$ is obtained by adding noise $\epsilon$ to semantic output $z_H^{sem}$ by ${z_t} = \sqrt {{{\bar \alpha }_t}} \cdot {z_H^{sem}} + \sqrt {1 - {{\bar \alpha }_t}}  \cdot \epsilon $, where the noise $\epsilon$ is sampled from $\epsilon \sim \mathcal{N}(0, I)$.
$w_t = \frac{1 - {{\bar \alpha }_t}}{\sqrt {{{\bar \alpha }_t}}} \cdot \frac{CS}{\|f\left( {z_t},\epsilon_{real}^{\lambda_{c\!f\!g}} \right) - z_{H}^{sem}\|_1}$ is the timestep-dependent scalar weight introduced in DMD \cite{dmd} to improve the training dynamics, where $S$ is the number of spatial locations, $C$ is the number of channels, and $f(\cdot)$ is the function defined in Eq. (1) of the main paper, $\epsilon_{real}^{{\lambda _{c\!f\!g}}}$ denotes the pre-trained SD output with the classifier-free guidance (CFG) term $\epsilon_{real}^{cls}$ defined in Eq. (6) of the main paper.

Following \cite{dmd, osediff}, we calculate the distribution matching gradient within the latent space instead of the noise domain. Therefore, Eq. (\ref{eq:SD_csd1}) can be written as:
\begin{equation}
    \begin{aligned}
       \nabla \ell _{CSD}^{{\lambda _{c\!f\!g}}}
       & = {\mathbb{E}_{t,\epsilon,{z_t},c}} \left[\frac{{w_t\sqrt {{{\bar \alpha }_t}} }}{{1 - {{\bar \alpha }_t}}} \left( {f\left( {{z_t},{\epsilon_{real}}({z_t},t)} \right) - f\left( {{z_t},{\epsilon_{real}}({z_t},t,c)} \right)} \right) \frac{{\partial {z_H^{sem}}}}{{\partial {\theta _{P\!i\!S\!A}}}}\right]\!,
    \end{aligned}
    \tag{S2}
    \label{eq:SD_csd2}
\end{equation}

We apply the CFG component to $f\left( {{z_t},{\epsilon_{real}}({z_t},t,c)} \right)$ in Eq. (\ref{eq:SD_csd2}) and merge the timestep-related weights to obtain Eq. (\ref{eq:SD_csd4}), which is the form of Eq. (5) in the main paper.

\begin{equation}
    \begin{aligned}
       \nabla \ell _{CSD}^{{\lambda _{c\!f\!g}}}
       & = {\mathbb{E}_{t,\epsilon,{z_t},c}} \left[\frac{CS}{\|f\left({z_t},\epsilon_{real}^{\lambda_{c\!f\!g}} \right) - z_{H}^{sem}\|_1} \left( {f\left( {{z_t},{\epsilon_{real}}}(z_t,t) \right) - f\left( {z_t},\epsilon_{real}^{\lambda_{c\!f\!g}} \right)} \right) \frac{{\partial {z_H^{sem}}}}{{\partial {\theta _{P\!i\!S\!A}}}}\right]\!.
    \end{aligned}
    \tag{S4}
    \label{eq:SD_csd4}
\end{equation}

\section{Comparisons with GAN-based methods}
\label{sec:GAN}
We compare PiSA-SR with three representative GAN-based SR methods: RealESRGAN \cite{realesrgan}, BSRGAN \cite{bsrgan}, and LDL \cite{LDL}. The quantitative results are presented in Table \ref{tab:comparison_gan}. PiSA-SR achieves the best performance on no-reference metrics (NIQE \cite{niqe}, CLIPIQA \cite{clipiqa}, MUSIQ \cite{musiq}, and MANIQA \cite{maniqa}) across the three benchmark datasets \cite{div2k, realsr, drealsr}, due to the enhanced generative capacity of the pre-trained SD model.

\begin{figure*}
	\centering 
    	\includegraphics[scale=1.3]{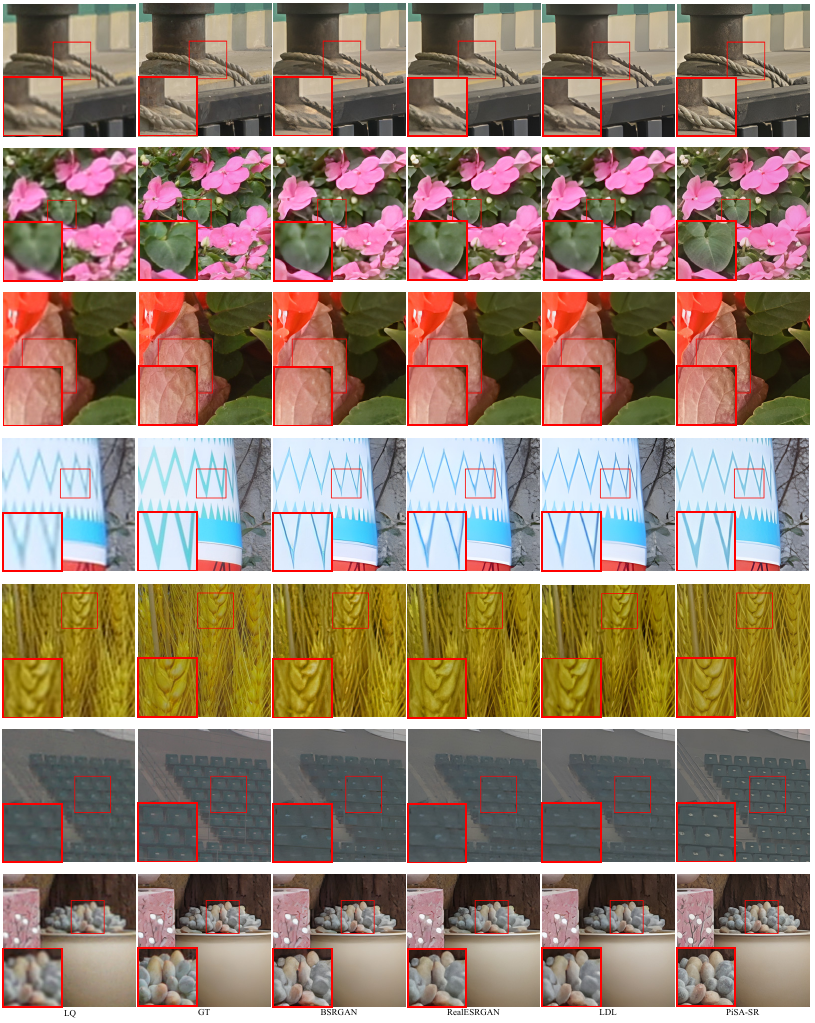}
     \vspace{-1em}
    	\caption{Visual comparisons between PiSA-SR and different GAN-based SR methods. Please zoom in for a better view.}
    	\label{fig:comparison_gan}
 \vspace{-1em}
\end{figure*}

In addition, PiSA-SR demonstrates competitive results on reference-based metrics (\eg, LPIPS \cite{lpips} and DISTS \cite{dists}), demonstrating a strong balance between perceptual quality and content fidelity. Fig. \ref{fig:comparison_gan} provides visual comparisons between PiSA-SR and the GAN-based methods. PiSA-SR generates more realistic details from LQ images, such as the regular textures of rope (see the first group) and the intricate textures of leaves (see the second and third groups).

\input{table/supp_comparison_GAN}

\begin{figure*}
	\centering 
    	\includegraphics[scale=0.6]{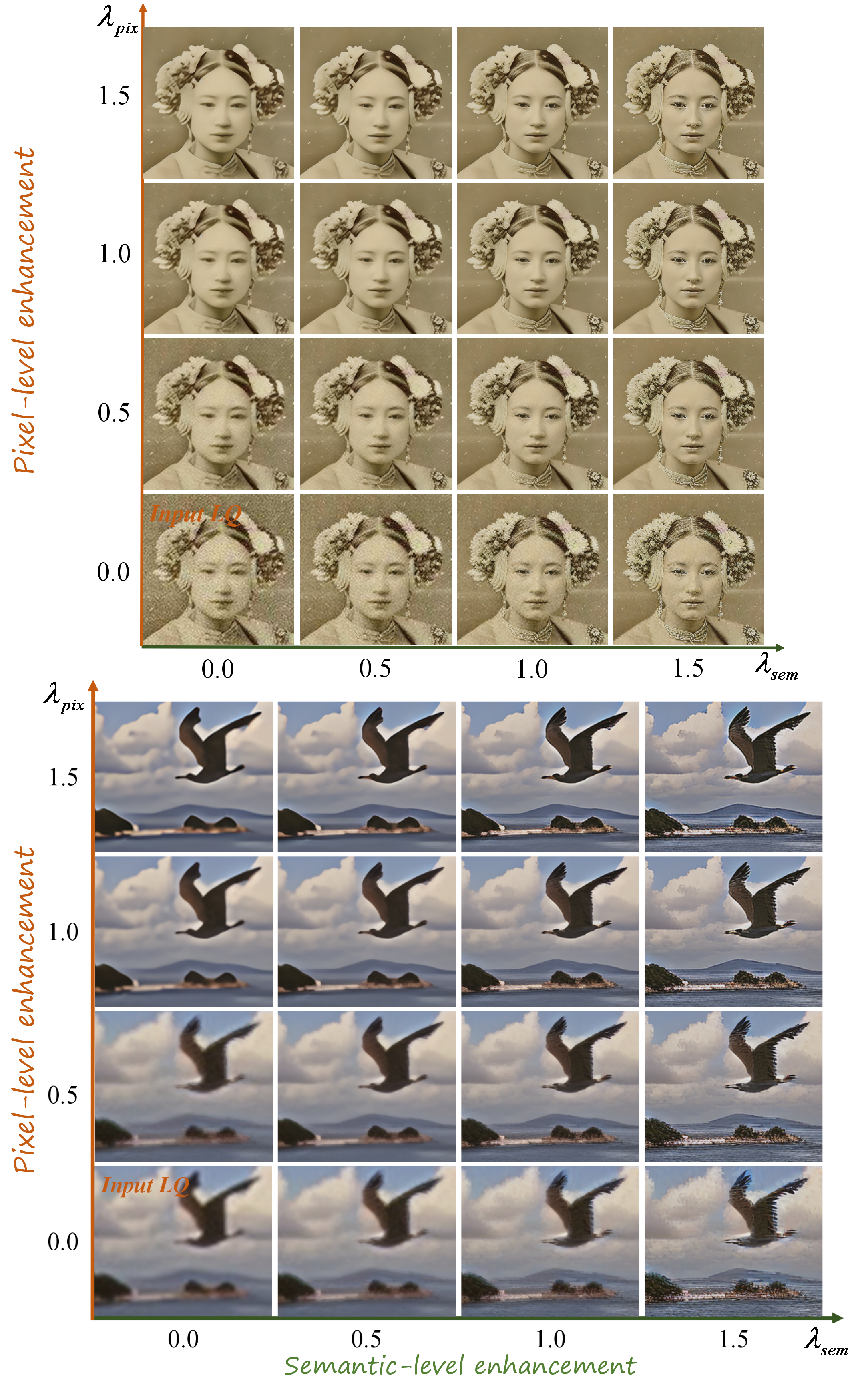}
     \vspace{-1em}
    	\caption{Visual examples of PiSA-SR with different pixel-semantic scales. The horizontal and vertical axes indicate the semantic-level and pixel-level enhancement scales, respectively.}
    	\label{fig:comparison_adj}
 \vspace{-1em}
\end{figure*}

\section{User study}
To further validate the effectiveness of the proposed adjustable SR method, we conducted a user study. Each participant was presented with a series of LQ images and their corresponding HQ outputs restored by our method. Specifically, for each LQ input, we applied a fixed pixel-level guidance factor of 1.0, ensuring consistency in pixel-level enhancement. Then, two HQ images were restored using two varying semantic guidance factors of 0.6 and 1.2, respectively.
During the study, participants were asked to select one of the two generated HQ images with superior semantic quality. The selection was considered positive if the participant chose the image generated with a semantic guidance factor of 1.2, demonstrating that our method effectively enhances the semantic quality of the output by increasing the semantic guidance factor.

In total, 10 participants provided 500 votes on 50 different LQ images. The results showed a positive selection rate of 98\%, which strongly supports the effectiveness of our approach. Therefore, increasing the semantic guidance factor can lead to noticeable improvements in semantic quality, making the generated images more visually appealing and semantically faithful to the input structure.

\section{More visual examples on adjustable SR}
\label{sec:Adjustable}

In Fig. \ref{fig:comparison_adj}, we provide additional visual examples on adjustable SR. The horizontal and vertical axes represent the enhancement scales for semantic and pixel levels, respectively.
Increasing pixel-level enhancement gradually reduces degradation and sharpens edges, but excessive enhancement can blur finer details (\eg, the lady’s hair accessory in the first group of Fig. \ref{fig:comparison_adj}). On the other hand, raising the semantic-level enhancement improves overall scene details (\eg, the seagull's wings, sea ripples, and island trees in the second group of Fig. \ref{fig:comparison_adj}), but may introduce artifacts when over-enhanced.

\begin{figure*}
	\centering 
    	\includegraphics[scale=1.5]{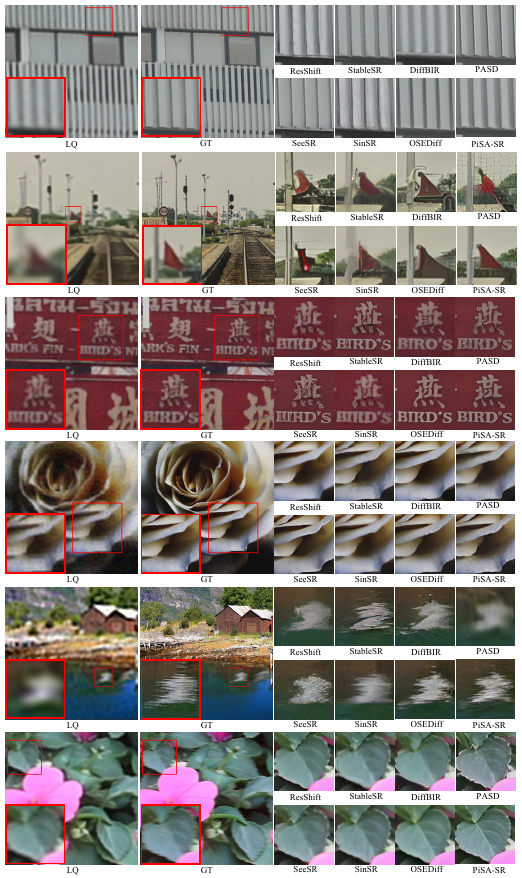}
     \vspace{-1em}
    	\caption{Visual comparisons of different DM-based SR methods. Please zoom in for a better view.}
    	\label{fig:comparison_dm}
 \vspace{-1em}
\end{figure*}

\section{More visual comparisons with DM-based SR methods}
\label{sec:DM}
We provide more visual comparisons of DM-based SR methods in Fig. \ref{fig:comparison_dm}. PiSA-SR surpasses other methods by reconstructing more accurate structures (\eg, the red flag's structure in the second group) and producing more realistic details (\eg, the flower's texture in the fourth group and the water ripples in the fifth group of Fig. \ref{fig:comparison_dm}).

\section{Ablation studies}



\subsection{Effectiveness of the dual-LoRA training}
To validate the effectiveness of the proposed dual-LoRA training approach, we conduct three experiments, as shown in Table \ref{tab:ablation_pisa}: optimizing only the pixel-level LoRA with $\ell_2$ loss (V1), optimizing only the semantic-level LoRA with LPIPS and CSD losses (V2), and applying the proposed dual-LoRA training (PiSA-SR). The performance is evaluated on the RealSR test dataset \cite{realsr}.
In V1, the no-reference-based metrics are relatively poor, suggesting that the results lack finer details.
In V2, while only optimizing with semantic losses, the reference-based metrics decline, introducing weakened fidelity and unnatural details.
In contrast, the proposed dual-LoRA training (PiSA-SR) strikes a better balance between fidelity and perceptual quality. It preserves fine pixel-level details in LQ while enhancing semantic details, resulting in more visually pleasing and natural-looking results.

\subsection{Comparison between CFG and the proposed semantic-level guidance}
CFG \cite{cfg} is a commonly used strategy to enhance semantic information in text-to-image (T2I) tasks \cite{stablediffusion} and multi-step DM-based SR methods \cite{pasd, seesr}. However, it becomes ineffective in one-step DMs distilled from multi-step DMs \cite{dmd, osediff}.
As discussed in Sec. 4.2 of the main paper, our proposed semantic-level guidance offers an alternative for one-step DM-based SR methods to semantics enhancement. To further demonstrate its effectiveness, we compare the proposed semantic-level guidance with CFG on the RealSR \cite{realsr} test dataset in Table \ref{tab:ablation_cfg}.
Unlike in multi-step DM-based SR methods, increasing the CFG scale in one-step DMs does not improve no-reference metrics (CLIPIQA \cite{clipiqa}, MUSIQ \cite{musiq}) and can even degrade both reference- and no-reference-based metrics (\eg, LPIPS \cite{lpips} and CLIPIQA \cite{clipiqa}) at higher scales (\eg, $\lambda_{c\!f\!g}=3.0$). In contrast, adjusting the proposed semantic-level guidance effectively boosts semantic information in the restored image (\eg, changing $\lambda_{sem}$ from 1.0 to 1.2 can increase the MUSIQ metric from 70.15 to 71.60). Additionally, applying CFG requires more inference time than PiSA-SR, since it needs to extract text prompts from LQ to generate conditional outputs.

\input{table/ablation}

\subsection{Impact of LoRA rank}
In our default setting, the ranks of both pixel-level and semantic-level LoRAs are set to 4.
We conduct experiments by fixing one LoRA rank at 4 and varying the other to observe how the performance on the RealSR \cite{realsr} test dataset changes. The results are shown in Table \ref{tab:ablation_rank}. Increasing the semantic-level LoRA rank enhances semantic details, as reflected in the CLIPIQA and MUSIQ metrics. However, this comes at the cost of image fidelity, resulting in lower reference-based scores (\eg, LPIPS).
A similar pattern can be observed for pixel-level LoRA. Raising its rank improves fidelity, as shown by the SSIM and LPIPS, but reduces some image details, leading to a drop in no-reference metrics (\eg, MUSIQ).
Overall, increasing the LoRA rank for one task (pixel-level or semantic-level enhancement) improves its performance but deteriorates the other. This is due to the inherent conflict between these two tasks \cite{pdtradeoff}. We choose a rank of 4 for both LoRA modules as it offers a good balance.

\subsection{Impact of the input timestep}
In one-step DM-based SR methods, the input timestep must be set in advance. By default, we use a value of 1.
Table \ref{tab:ablation_t} presents the performance of different timestep configurations on the RealSR \cite{realsr} test dataset. Generally speaking, increasing the input timestep enhances the semantic details in the restored images but worsens the fidelity performance. This is because the pre-trained models with larger input timestep have stronger denoising capabilities, which improve the generation capability but reduce the ability to preserve the input structures.
In PiSA-SR, we set the input timestep to 1 to retain the LQ image information as much as possible.

\begin{figure*}
            \centering 
    	\includegraphics[scale=0.65]{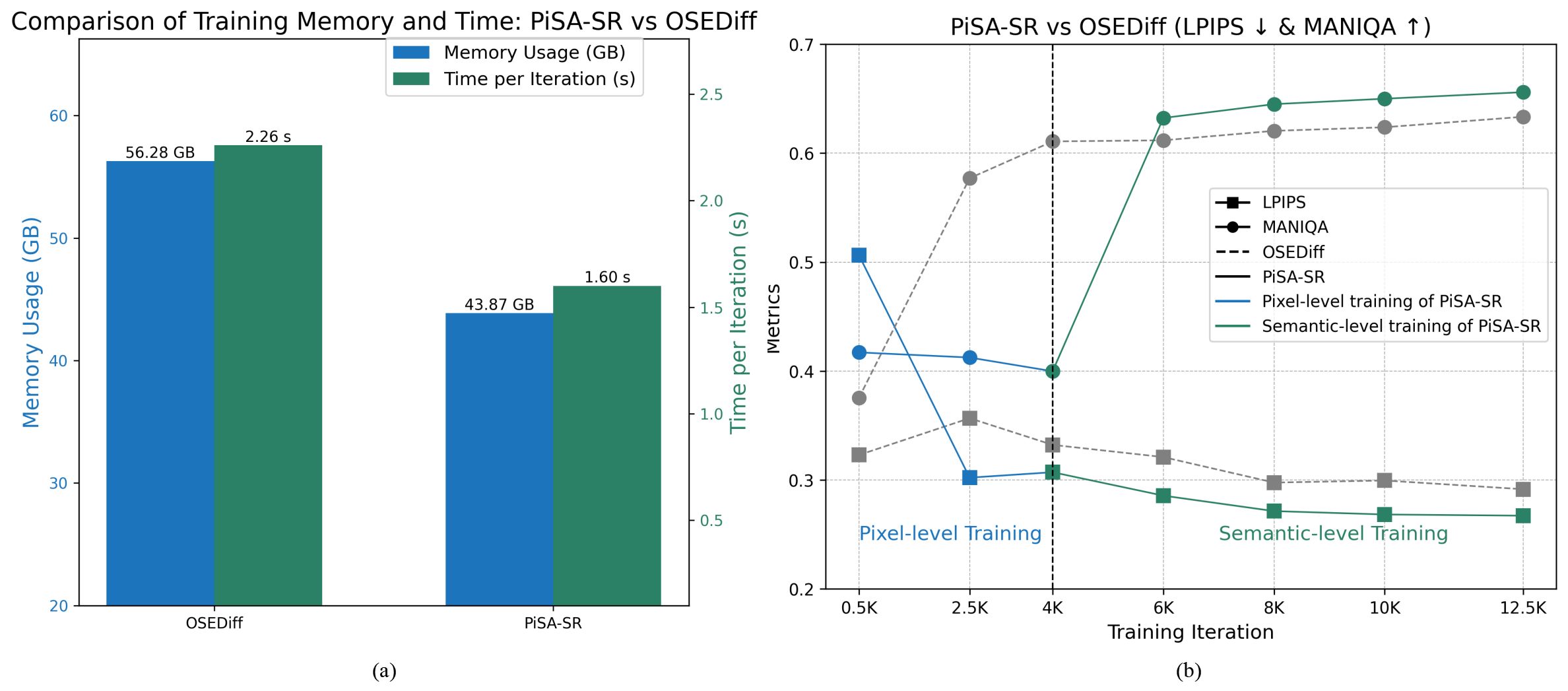}
    	\caption{Comparison between OSEDiff and PiSA-SR: (a) Training memory consumption and time per iteration, and (b) performance (LPIPS \cite{lpips} and MANIQA \cite{maniqa}) on the RealSR test dataset across training iterations. Experiments are conducted on a single NVIDIA A100 80G GPU with a batch size of 4.}
    	\label{fig:comparison_osediff1}
\end{figure*}

\begin{figure*}
            \centering 
    	\includegraphics[scale=0.75]{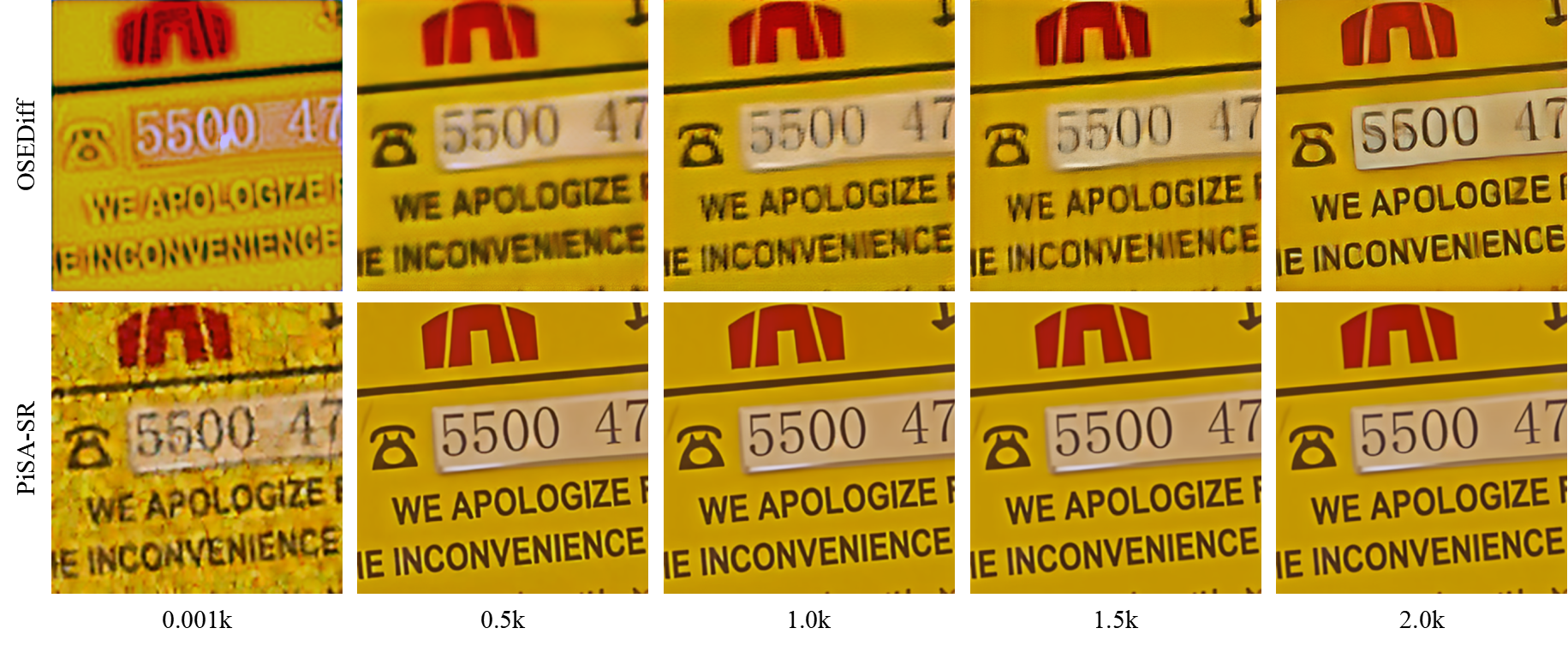}
    	\caption{Visual comparisons of SR results from OSEDiff and PiSA-SR across 1 to 2000 training iterations.}
    	\label{fig:comparison_osediff2}
\end{figure*}

\subsection{Comparison with OSEDiff}
In Sec. 4.3 of the main paper, we present both quantitative and qualitative comparisons between PiSA-SR and OSEDiff \cite{osediff}. Here, we compare them in three additional aspects: training memory usage, training time per iteration, and convergence speed.
PiSA-SR uses the CSD loss \cite{csd} for semantic enhancement, while OSEDiff employs the VSD loss \cite{vsd} for distillation. As discussed in Sec. 3.3 of the main paper, optimizing with the CSD loss provides richer semantic information for the SR model than VSD. Furthermore, CSD does not require bi-level optimization, significantly reducing training memory usage and training time per iteration, as shown in Fig. \ref{fig:comparison_osediff1} (a).
PiSA-SR formulates the SR task by learning the residual between LQ and HQ latent features (see Sec. 3.1 of the main paper), accelerating training convergence as evidenced in Fig. \ref{fig:comparison_osediff2}, where PiSA-SR restores clearer text faster than OSEDiff during the first 2000 training iterations.

We also compare the behaviors of PiSA-SR and OSEDiff over the training process, measured by the reference-based LPIPS and no-reference-based MANIQA metrics, as shown in Fig. \ref{fig:comparison_osediff1} (b). In the first 4K iterations, PiSA-SR focuses on pixel-level LoRA optimization, improving LPIPS but slightly worsening MANIQA. Because OSEDiff applies LPIPS and VSD losses throughout the training process, it initially outperforms PiSA-SR in MANIQA.
After 4K iterations, PiSA-SR begins optimizing the semantic-level LoRA. Thanks to its faster convergence, PiSA-SR outperforms OSEDiff in LPIPS and MANIQA between 4K and 6K iterations. Beyond 6K iterations, both methods continue to improve, but PiSA-SR consistently demonstrates superior performance throughout the training process.
Overall, PiSA-SR achieves faster training speeds and lower memory consumption than OSEDiff, despite utilizing pixel-level and semantic-level optimizations.

%% file: table/supp_comparison_GAN.tex
\begin{table*}\tiny
           \centering

 \caption{
        Quantitative comparison among the state-of-the-art GAN-based SR methods on synthetic and real-world test datasets. The best results are highlighted in {\color[HTML]{FF0000} \textbf{red}}.}
 
   \resizebox{\linewidth}{!}{
\begin{tabular}{c|c|ccccccccc}
\hline
Datasets                   & Methods                                                  & PSNR↑&SSIM↑& LPIPS↓& DISTS↓& FID↓                                    & NIQE↓ & CLIPIQA↑ & MUSIQ↑                                  & MANIQA↑                                                   \\

\hline
                           & RealESRGAN                                            & {24.29}&{\color[HTML]{FF0000} \textbf{0.6371}}&0.3112& 0.2141& 37.64&4.68& 0.5277& 61.06&         0.5501\\
                           
                           & BSRGAN                                         & {\color[HTML]{FF0000} \textbf{24.58}}& 0.6269&0.3351& 0.2275& 44.23& 4.75& 0.5071& 61.20&         0.5247\\
                           
                           & LDL                                              & 23.83& 0.6344&0.3256& 0.2227& 42.29& 4.85& 0.5180& 60.04&         0.5350\\

\rowcolor{lightpink}
 \multirow{-4}{*}{DIV2K}& PiSA-SR-S1& 23.87& 0.6058& {\color[HTML]{FF0000} \textbf{0.2823}}& {\color[HTML]{FF0000} \textbf{0.1934}}& {\color[HTML]{FF0000} \textbf{25.07}}& {\color[HTML]{FF0000} \textbf{4.55}}& {\color[HTML]{FF0000} \textbf{0.6927}}& {\color[HTML]{FF0000} \textbf{69.68}}& {\color[HTML]{FF0000} \textbf{0.6400}}\\
\hline 
                           &RealESRGAN                                     & {\color[HTML]{FF0000} \textbf{25.69}} & 0.7616 & 0.2727 & 0.2063 & 135.18 & 5.83 & 0.4449    & 60.18                               &        0.5487                \\
                           & BSRGAN                                               & 26.39& {\color[HTML]{FF0000} \textbf{0.7654}} & {\color[HTML]{FF0000} \textbf{0.2670}} & 0.2121 & 141.28 & 5.66  & 0.5001       &  63.21             &         0.5399      \\
                           & LDL                                              & 25.28 & 0.7567 & 0.2766 & 0.2121 & 142.71 & 6.00 & 0.4477     & 60.82                                &         0.5485                 \\

\rowcolor{lightpink}
\multirow{-4}{*}{RealSR}  & PiSA-SR-S1 & 25.50& {\color[HTML]{FF0000} \textbf{0.7417}}& 0.2672& {\color[HTML]{FF0000} \textbf{0.2044}}& {\color[HTML]{FF0000} \textbf{124.09}}& {\color[HTML]{FF0000} \textbf{5.50}}& {\color[HTML]{FF0000} \textbf{0.6702}}& {\color[HTML]{FF0000} \textbf{70.15}}& {\color[HTML]{FF0000} \textbf{0.6560}}\\
  \hline 
                           & RealESRGAN                                               & 28.64 & 0.8053 & 0.2847 & {\color[HTML]{FF0000} \textbf{0.2089}} & 147.62 & 6.69   & 0.4422        &54.18                     &        0.4907                 \\
                           & BSRGAN                                              & {\color[HTML]{FF0000} \textbf{28.75}} & 0.8031 & 0.2883 & 0.2142 & 155.63 & 6.52     & 0.4915        & 57.14                            &    0.4878            \\
                           & LDL                                           & 28.21 & {\color[HTML]{FF0000} \textbf{0.8126}} & {\color[HTML]{FF0000} \textbf{0.2815}} & 0.2132 & 155.53 & 7.13 & 0.4310    & 53.85                               &        0.4914                \\
                           
\rowcolor{lightpink}
\multirow{-4}{*}{DrealSR}& PiSA-SR-S1&28.31 & 0.7804 & 0.2960& 0.2169& {\color[HTML]{FF0000} \textbf{130.61}}& {\color[HTML]{FF0000} \textbf{6.20}}&{\color[HTML]{FF0000} \textbf{0.6970}} & {\color[HTML]{FF0000} \textbf{66.11}}& {\color[HTML]{FF0000} \textbf{0.6156}}\\ \hline
\end{tabular}
}
\label{tab:comparison_gan}
\end{table*}

%% file: table/ablation.tex

\begin{table}\footnotesize

\caption{Ablation studies on the dual-LoRA training approach on RealSR dataset.}
\centering
\begin{tabular}{ccc|cccccc}
\hline
 Methods & Pixel-level LoRA & Semantic-level LoRA & PSNR↑ & SSIM↑ & LPIPS↓ & CLIPIQA↑ & MANIQA↑ & MUSIQ↑ \\ \hline
          V1& \color{blue}{$\checkmark$}& $\color{red}{\times}$&  27.28 & 0.7975& 0.3090  &  0.3130 & 0.3995 & 49.02 \\
          V2&     $\color{red}{\times}$& \color{blue}{$\checkmark$}&  24.13 & 0.7290 &  0.2803   &0.6711 & 0.6614  &  70.69 \\ \hline
          \rowcolor{lightpink}
          PiSA-SR&     \color{blue}{$\checkmark$}&           \color{blue}{$\checkmark$}&   25.50 & 0.7417  &   0.2672    &   0.6702 & 0.6560&  70.15  \\ \hline
\end{tabular}
\label{tab:ablation_pisa}
\end{table}

\begin{table}\footnotesize
\caption{Ablation studies on pixel-level and semantic-level LoRA ranks on RealSR dataset.}
\centering
\begin{tabular}{ccc|cccccc}
\hline
 Methods & Pixel-level LoRA rank & Semantic-level LoRA rank & PSNR↑ &SSIM↑& LPIPS↓ & CLIPIQA↑ & MANIQA↑ & MUSIQ↑  \\ \hline
            \rowcolor{lightpink}
         PiSA-SR&    4&           4 &   25.50 & 0.7417  &   0.2672    &   0.6702 & 0.6560&  70.15 \\ \hline
          R1& 4& 8& 25.40 & 0.7401 & 0.2719  & 0.6774 & 0.6584& 69.93 \\
          R2&    4 &16 & 25.36&0.7398 &  0.2726  &0.6777  & 0.6603&70.15 \\ 
          R3&    4 &32 & 25.40 & 0.7394 &  0.2713 &0.6784 & 0.6634 &70.13 \\  \hline
          R4& 8& 4&  25.39 & 0.7422 &  0.2628  &  0.6663 & 0.6600& 69.86 \\
          R5&    16  & 4 &  25.54  & 0.7511&0.2624  &0.6603 & 0.6636& 69.77\\ 
          R6&    32 & 4 & 26.01 & 0.7565 & 0.2564  & 0.6318  & 0.6409 & 68.30\\  \hline

\end{tabular}
\label{tab:ablation_rank}
\end{table}

\begin{table}\footnotesize
\caption{Ablation studies on the input timestep on RealSR dataset.}
\centering
\begin{tabular}{cc|cccccc}
\hline
 Methods & Input timestep  & PSNR↑  &SSIM↑ & LPIPS↓ & CLIPIQA↑ & MANIQA↑ & MUSIQ↑ \\ \hline
            \rowcolor{lightpink}
         PiSA-SR&    1 &   25.50   & 0.7417 &  0.2672    &   0.6702 & 0.6560 &  70.15  \\ \hline
          I1& 250&  25.52& 0.7398 &  0.2685  &  0.6705 &0.6572& 70.05 \\
          I2&    500 & 25.42& 0.7383 &   0.2699  &0.6728  &0.6596 & 70.09 \\ 
          I3&    750 &  25.38 & 0.7376 &  0.2701  &0.6750  &0.6598&70.11 \\ 
          I4& 999 &   25.25& 0.7330 &  0.2707  &  0.6813&0.6628& 70.26 \\ \hline
          
\end{tabular}
\label{tab:ablation_t}
\end{table}

\begin{table}\footnotesize
\caption{Comparisons of CFG and the proposed semantic-level guidance on RealSR dataset.}
\centering
\begin{tabular}{cc|cccc|c}
\hline
 $\lambda_{cfg}$ & $\lambda_{sem}$ & PSNR↑ & LPIPS↓ & CLIPIQA↑ & MUSIQ↑ & Inference time(s)/Image\\ \hline
            \rowcolor{lightpink}
           1.0 & 1.0 & 25.50   &   0.2672    &   0.6702 &  70.15 & 0.09 \\ \hline
          1.2 & $\color{red}{\times}$ &  25.38 &  0.2684  &  0.6708 & 70.23 & 0.15\\
          1.5 &  $\color{red}{\times}$ & 25.30 &  0.2698  &0.6708  &70.29 & 0.15\\ 
          3.0 & $\color{red}{\times}$ & 24.61 &  0.2834  &0.6540  &70.14 & 0.15\\ \hline
          $\color{red}{\times}$ &  1.2 & 24.59 &   0.3000   &   0.7015 &   71.60 & 0.13 \\ 
          $\color{red}{\times}$ & 1.5 & 23.08 &   0.3541  &  0.6835 &   71.76 & 0.13 \\ \hline
          
\end{tabular}
\label{tab:ablation_cfg}
\end{table}


